\documentclass[lettersize,journal]{IEEEtran}
\usepackage{amsmath,amsfonts}
\usepackage{algorithmic}
\usepackage{algorithm}
\usepackage{array}
\usepackage[caption=false]{subfig}
\usepackage{textcomp}
\usepackage{stfloats}
\usepackage{url}
\usepackage{verbatim}
\usepackage{graphicx}

\usepackage[utf8]{inputenc} 
\usepackage[T1]{fontenc}    
\usepackage{url}            
\usepackage{booktabs}       
\usepackage{amsfonts}       
\usepackage{nicefrac}       
\usepackage{microtype}      
\usepackage{xcolor}         

\usepackage{makecell}
\usepackage{amsmath}
\usepackage{accents}
\usepackage{statex}
\usepackage[normalem]{ulem}
\usepackage{enumitem}
\usepackage{multirow}

\usepackage{changepage}
\usepackage{bm}
\usepackage{mathrsfs}

\usepackage{diagbox}
\usepackage{pdftexcmds}
\usepackage{catchfile}
\usepackage{ifluatex}
\usepackage{ifplatform}
\usepackage{threeparttable}

\usepackage{comment}
\usepackage{aecompl}
\usepackage{lipsum}   
\begin{document}

\title{One Fits All: General Mobility Trajectory Modeling via Masked Conditional Diffusion}

\author{Qingyue Long, Can Rong, Huandong Wang, Yong Li
\thanks{Manuscript received January 7, 2025; revised xxxx, 2025.}
\thanks{The authors are with the Department of Electronic Engineering, Beijing National Research Center for Information
Science and Technology (BNRist), Tsinghua University, Beijing 100084, China (e-mail: longqy21@mails.tsinghua.edu.cn; rc20@mails.tsinghua.edu.cn;
wanghuandong@tsinghua.edu.cn;
liyong07@tsinghua.edu.cn)
}
}

\markboth{IEEE TRANSACTIONS ON KNOWLEDGE AND DATA ENGINEERING,~VOL.~36, NO.~12, JANUARY~2025}%
{Shell \MakeLowercase{\textit{et al.}}: One Fits All: General Mobility Trajectory Modeling via Masked Conditional Diffusion}

\IEEEpubid{0000--0000~\copyright~2025 IEEE}

\maketitle

\begin{abstract}
Trajectory data play a crucial role in many applications, ranging from 
network optimization to urban planning.
Existing studies on trajectory data are task-specific, and their applicability is limited to the specific tasks on which they have been trained, such as generation, recovery, or prediction.
However, the potential of a unified model has not yet been fully explored in trajectory modeling. 
Although various trajectory tasks differ in inputs, outputs, objectives, and conditions, they share common mobility patterns.
Based on these common patterns, we can construct a general framework that enables a single model to address different tasks.
However, building a trajectory task-general framework faces two critical challenges: 1) the diversity in the formats of different tasks and 2) the complexity of the conditions imposed on different tasks.
In this work, we propose a general trajectory modeling framework via masked conditional diffusion (named \textbf{GenMove}). 
Specifically, we utilize mask conditions to unify
diverse formats. 
To adapt to complex conditions associated with different tasks, we utilize historical trajectory data to obtain contextual trajectory embeddings, which include rich contexts such as spatiotemporal characteristics and user preferences. 
Integrating the contextual trajectory embedding into diffusion models through a classifier-free guidance approach allows the model to flexibly adjust its outputs based on different conditions.
Extensive experiments on mainstream tasks demonstrate that our model significantly outperforms state-of-the-art baselines, with the highest performance improvement exceeding 13\% in generation tasks. 
\end{abstract}


\section{Introduction}

Trajectory data record human movement and reflect the patterns of human activities, supporting various applications, such as urban planning, traffic control, and emergency response~\cite{li2025generative, wang2021public, yabe2024enhancing}. Given the rich information embedded in trajectory data, researchers have developed methods to solve tasks like trajectory generation, recovery, and prediction.
Although numerous studies have addressed various trajectory tasks, most existing works still rely on models specifically designed for a single task. For example, in trajectory prediction, researchers typically develop specialized models that use historical trajectory data to predict future locations~\cite{feng2018deepmove, long2024universal}. In trajectory generation, deep generative models are used to generate synthetic trajectory data that conform to realistic mobility patterns~\cite{jiang2016timegeo, cao2021generating}. For trajectory recovery, the common approach involves using statistical interpolation or deep learning techniques to address missing data in trajectories~\cite{elshrif2022network,xia2021attnmove}. 
While these methods excel in specific tasks, they are limited by a lack of flexibility and generality. 

\begin{figure}[t]
\centering
\includegraphics[width=0.9\linewidth]{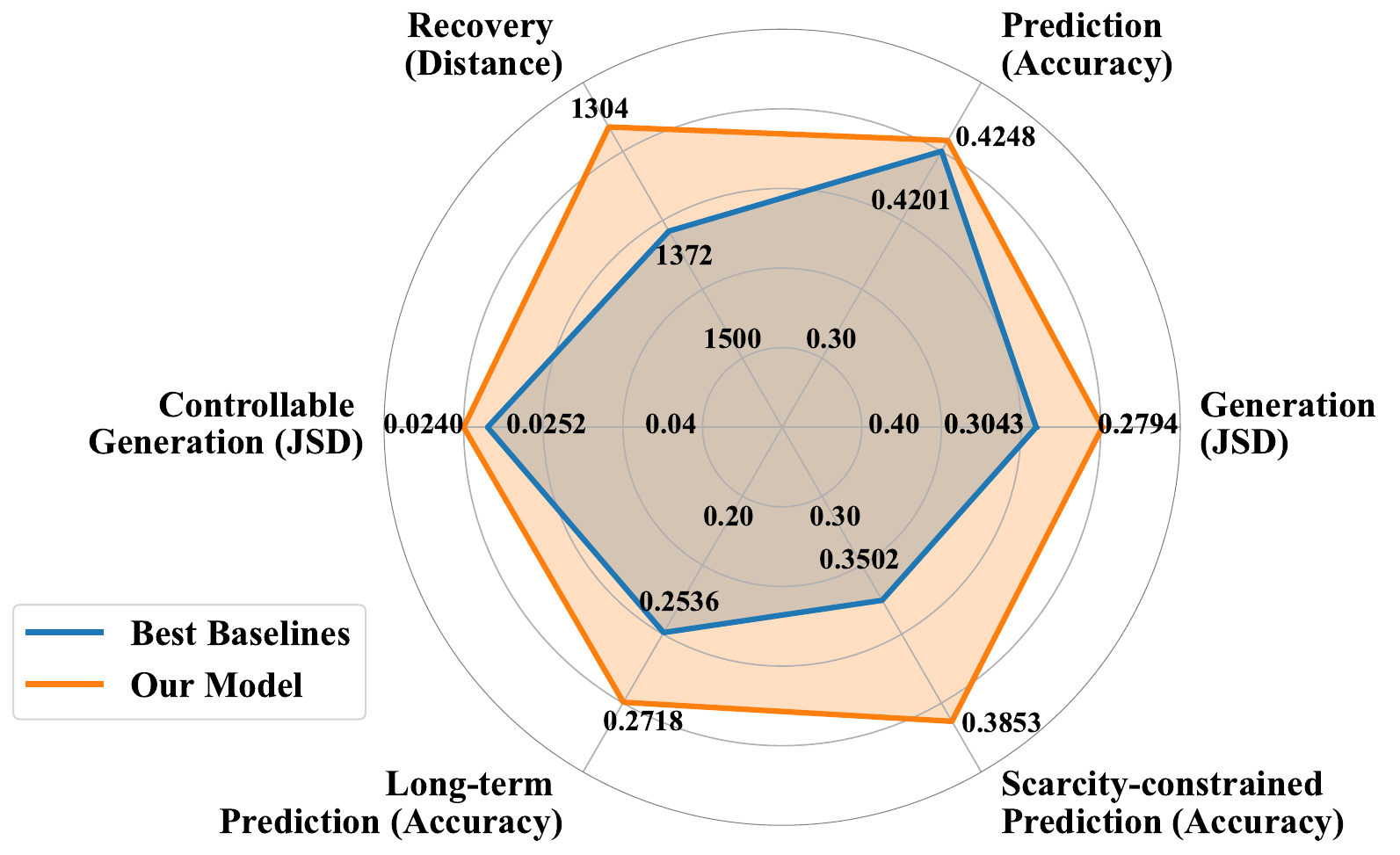}
\caption{Performance of GenMove on ISP dataset.}
\vspace{-0.4cm} 
\label{fig:Performance}
\end{figure}

In fact, different trajectory tasks essentially rely on common human mobility patterns, such as periodicity, sequentiality, and circadian rhythms. Leveraging these shared patterns across various tasks within the same framework can significantly enhance the performance of each task.
On the other hand, different application scenarios have diverse conditions for trajectory tasks. For example, in traffic management, short-term trajectory prediction is needed to optimize real-time traffic flow~\cite{xu2024representation}. In contrast, urban planning relies on long-term trajectory prediction to design the layout of future infrastructure~\cite{rossi2021vehicle}. When limited data is available due to privacy protections~\cite{chen2024advancements}, unconditional trajectory generation can provide a substantial amount of generated data~\cite{zhu2023diffusion,wang2023pategail}. Moreover, controllable trajectory generation is necessary for personalized advertising or travel recommendations to meet user needs preciselys~\cite{dai2015personalized,xing2017personalized}. 
Therefore, it is necessary to develop a general framework that can flexibly adapt to various application scenarios.

\IEEEpubidadjcol
There are remarkable achievements of general frameworks in computer vision and natural language processing~\cite{long2022vision,xie2023towards}, such as GPT, which can perform various tasks, including text translation, question answering, and image generation from text~\cite{achiam2023gpt}. Inspired by these developments, we naturally pose a new research question: \textit{Can we address multiple tasks with a general framework for mobility modeling?}
However, the task-general mobility modeling framework is still an open problem with the following challenges:
\begin{itemize}[leftmargin=*]
\item \textbf {Diversity of different task formats.} 
Different tasks are naturally described in different forms, which differ in their inputs, outputs, and objectives.
For example, trajectory prediction typically relies on past trajectory points, whereas trajectory recovery demands both past and future trajectory points. More importantly, these tasks focus on different dependencies between trajectory points. 
Therefore, it's challenging to model these diverse tasks within a general framework.

\item \textbf{Complexity of the conditions imposed on different tasks.} Different trajectory tasks face varying conditions, which may be determined by external settings or inherent to the tasks. For example, unlike unconditional trajectory generation, controllable generation relies on setting conditions such as radius of gyration and speeds. Moreover, there are inherent differences in conditions between different tasks. For example, trajectory prediction requires modeling the historical patterns of specific users. In contrast, trajectory generation does not focus on specific users but rather captures the mobility patterns of the entire group. 
Thus, developing a general framework that can adapt to complex conditions is crucial but challenging.
\end{itemize}

To address the above challenges, we propose a general mobility modeling framework via masked conditional diffusion named \textbf{GenMove}.
Firstly, we design a mask condition module to unify multiple tasks into a common format of task objectives and conditional observations. We utilize various masking strategies, including task-specific and pattern-general masks. Task-specific masks precisely shield different trajectory parts, effectively adapting to the specific objectives of different tasks. Additionally, pattern-general masks model common mobility patterns across various tasks. The mask is used as a condition for the diffusion model, enabling a single model to address multiple mobility modeling tasks.
Secondly, we obtain contextual trajectory embedding from historical data to provide essential context for specific tasks. Then, we apply the contextual trajectory embedding to diffusion models using a classifier-free guidance approach, which enables the model to flexibly adjust its outputs based on different conditions.
Experimentally, as shown in Figure~\ref{fig:Performance}, GenMove achieves state-of-the-art performance on multiple tasks.
Overall, the contributions of our paper can be summarized as follows:
\begin{itemize}[leftmargin=*]
\item We unify mobility trajectory modeling, aiming to explore the vast potential of a general model capable of handling various trajectory tasks.
\item We propose a general trajectory modeling framework via masked conditional diffusion. Specifically, well-designed mask conditions unify diverse formats of different tasks. Then, it employs a classifier-free guidance approach to incorporate context information from the trajectory, addressing the complex conditions associated with different tasks.
\item We conduct extensive experiments on six trajectory-related tasks using two real-world datasets. 
The results demonstrate that our method outperforms state-of-the-art models across six tasks, with performance enhancements exceeding 13\% in trajectory generation.
\end{itemize}

\section{Related Work}
\paragraph{Trajectory-based Tasks}
In practical applications, there exist typical trajectory-based tasks such as trajectory generation, recovery, and prediction~\cite{feng2020learning, sun2021periodicmove,qingyue2024privacy}. The objectives and methods of different tasks vary significantly.
Trajectory generation depends on learning the distribution of real trajectories to generate new trajectory data. For example, the EPR model uses the patterns of exploring new locations and returning familiar ones to generate trajectories~\cite{song2010modelling}.
TrajGen uses GAN to learn the spatial distribution of trajectory locations in the original data~\cite{cao2021generating}.
Trajectory prediction uses historical trajectory information to predict the future location. For example, TrImpute relies on crowd wisdom to guide trajectory recovery using the spatial relationships of neighboring GPS points~\cite{elshrif2022network}. AttnMove utilizes various intra-trajectory and inter-trajectory attention mechanisms to better recover the mobility regularity of users~\cite{xia2021attnmove}.
Trajectory recovery focuses on reconstructing missing parts of trajectory data. For example, MobTCast uses a transformer-based context-aware network for mobility prediction~\cite{xue2021mobtcast}.
STAN uses non-contiguous but functionally similar visited points to
predict the next location~\cite{luo2021stan}.
Each task requires different model architectures and training strategies because of the different objectives. 
In this work, we propose a general model framework to address multiple trajectory-related tasks through a single model.


\paragraph{Task-general Models}
Task-general Models utilize a single model to handle multiple tasks, eliminating the need for separate algorithms for each task. They have rapidly advanced in various fields such as computer vision, natural language processing, and time series analysis~\cite{achiam2023gpt, liu2024sora, garza2023timegpt}.
Task-general models are typically divided into two categories. The first category involves building models from scratch specifically for a particular domain. For example, TimesNet is designed with specialized backbone networks that capture intra-periodic and inter-periodic temporal variations, addressing multiple time series analysis tasks~\cite{wu2022timesnet}. TrajGDM uses a trajectory generation framework to capture common mobility patterns in trajectory datasets, accomplishing trajectory-related tasks~\cite{chu2024simulating}.
The second category involves leveraging pre-trained large language models (LLMs) to adapt to specific domains. 
Time-LLM is a methodology that utilizes LLMs for broad time series forecasting while leaving the backbone language models intact~\cite{jin2023time}.
UrbanGPT is a general spatiotemporal prediction framework that integrates spatiotemporal information into language models. It enables LLMs to deeply understand the complex connections between time and space~\cite{li2024urbangpt}.
In this work, we develop a task-general framework from scratch for various trajectory modeling tasks. It can handle more complex tasks, such as controllable generation and scarcity-constrained prediction, beyond simple trajectory tasks.

\section{Preliminaries}
\subsection{Problem Definition}
\paragraph{Definition 1: (Mobility Trajectory).} The mobility trajectory of user $u$ is a sequence of locations sampled at equal time intervals, denoted by $s_u=\{l_1,l_2,...,l_n\}$, where each location $l_i$ is represented as the form in latitude and longitude coordinates or a region ID.

\paragraph{Definition 2: (General Mobility Trajectory Modeling Problem).}
The objective of this problem is to utilize a single model to accomplish various trajectory-related tasks, such as trajectory generation, recovery, and prediction.
\begin{itemize}[leftmargin=*]
    \item \textit{(Trajectory Generation).} Given a set of real-world collected trajectories, $\mathcal{S} = \{s_1,s_2,...,s_m\}$, where each $s_u=\{l_1,l_2,...,l_n\}$ is a sequence of locations, the trajectory generation problem aims to develop a generative model $G$ capable of generating synthetic trajectories.
    
    \item \textit{(Trajectory Recovery).} Given the sparse trajectory of user $u$ with some missing values $s_u=\{l_1, \varnothing,l_3,..., \varnothing,l_n\}$, where $\varnothing$ represents the unobserved location at the corresponding time slot, the objective of the trajectory recovery task is to recover the missing locations.
    
    \item \textit{(Trajectory Prediction).} Given user $u$'s past trajectory $s_u=\{l_1,l_2,...,l_n\}$, the goal of the trajectory prediction task is to estimate the probability of their next location $l_{n+1}$.
\end{itemize}

\subsection{Conditional Denoising Diffusion Probabilistic Models}
\label{sec:PRELIMINARIES}
The expression $p_\theta(x_0):=\int p_\theta(x_{0:T}) dx_{1:T}$ describes diffusion models, a type of latent variable model where $x_1,..., x_T$ are latent vectors matching the dimensionality of the data $x_0 \sim q(x_0)$. In the diffusion probabilistic model, two Markov chains are utilized: one to incrementally introduce noise to the data, and the other to reverse this process and restore the original data. The following Markov chain defines the diffusion process:
\begin{equation}\label{equ:DDPM1}
q\left(\mathbf{x}_{1: T} \mid \mathbf{x}_{0}\right):=\prod_{t=1}^{T} q\left(\mathbf{x}_{t} \mid \mathbf{x}_{t-1}\right), 
\end{equation}
where $q\left(\mathbf{x}_{t} \mid \mathbf{x}_{t-1}\right):=\mathcal{N}\left(\sqrt{1-\beta_{t}} \mathbf{x}_{t-1}, \beta_{t} \mathbf{I}\right)$ and $\beta_t$ are modest positive constants that reflect the noise intensity. $x_t$ may be represented in closed form as $x_{t}=\sqrt{\alpha_{t}} x_{0}+\left(1-\alpha_{t}\right) \epsilon.$ for $\epsilon \sim \mathcal{N}(0, \mathbf{I})$, where $\alpha_{t}=\sum_{i=1}^{t}\left(1-\beta_{t}\right)$.

Next, we discuss the conditional extension~\cite{tashiro2021csdi}, where the modeled $x$ is conditioned on observations $c$. Estimating the actual conditional data distribution is the goal of the conditional denoising diffusion probabilistic model. With a model distribution, $q(x_0|c)$ $p_\theta(x_0|c)$.

Conversely, the conditional reverse process, which is represented by the Markov chain below, denoises $x_t$ to retrieve $x_0$:
\begin{equation}\label{equ:DDPM2}
p_{\theta}\left(\mathbf{x}_{0: T}|c\right):=p\left(\mathbf{x}_{T}\right) \prod_{t=1}^{T} p_{\theta}\left(\mathbf{x}_{t-1} \mid \mathbf{x}_{t},c\right),
\end{equation}
where $\mathbf{x}_{T} \sim \mathcal{N}(\mathbf{0}, \mathbf{I})$. Again, $p_\theta(x_{t-1}|x_t,c)$ is assumed as normally distributed with learnable parameters as follows:
\begin{equation}\label{equ:DDPM3}
p_{\theta}\left(\mathbf{x}_{t-1} \mid \mathbf{x}_{t},c\right):=\mathcal{N}\left(\mathbf{x}_{t-1} ; \boldsymbol{\mu}_{\theta}\left(\mathbf{x}_{t}, t|c\right), \sigma_{\theta}\left(\mathbf{x}_{t}, t|c\right) \mathbf{I}\right).
\end{equation}
According to Ho et al.~\cite{ho2020denoising}, the unconditional denoising diffusion model takes into account the particular parameterization of $p_\theta(x_{t-1}|x_t)$ as follows:
\begin{equation}\label{equ:DDPM4}
\begin{cases}
\boldsymbol{\mu}_{\theta}\left(\mathbf{x}_{t}, t\right)=\frac{1}{\alpha_{t}}\left(\mathbf{x}_{t}-\frac{\beta_{t}}{\sqrt{1-\alpha_{t}}} \boldsymbol{\epsilon}_{\theta}\left(\mathbf{x}_{t}, t\right)\right), \\
\sigma_{\theta}\left(\mathbf{x}_{t}, t\right)=\tilde{\beta}_{t}^{1 / 2} \quad \text{where} \quad \tilde{\beta}_{t}=\left\{
\begin{array}{ll}
\frac{1-\alpha_{t-1}}{1-\alpha_{t}} \beta_{t} & t>1 \\
\beta_{1} & t=1
\end{array}\right.
\end{cases}
\end{equation}
where $\epsilon_\theta$ is a trainable denoising function. In Eq.~\ref{equ:DDPM4}, we represent $\boldsymbol{\mu}_{\theta}\left(\mathbf{x}_{t}, t\right)$ and $\sigma_{\theta}\left(\mathbf{x}_{t}, t\right)$ as $\mu^{\mathrm{DDPM}}\left(\mathbf{x}_{t}, t, \boldsymbol{\epsilon}_{\theta}\left(\mathbf{x}_{t}, t\right)\right)$ and $\sigma^{\mathrm{DDPM}}\left(\mathbf{x}_{t}, t\right)$.

Expanding upon the unconditional diffusion model, we derive a conditional denoising function $\epsilon_\theta$ that accepts the condition $c$ as input. Next, we examine the given parameterization with $\epsilon_\theta$:
\begin{equation}\label{equ:DDPM5}
\begin{cases}
\boldsymbol{\mu}_{\theta}\left(\mathbf{x}_{t}, t \mid c\right) = \boldsymbol{\mu}^{\mathrm{DDPM}}\left(\mathbf{x}_{t}, t, \boldsymbol{\epsilon}_{\theta}\left(\mathbf{x}_{t}, t \mid c\right)\right), \\
\sigma_{\theta}\left(\mathbf{x}_{t}, t \mid c\right) = \sigma^{\mathrm{DDPM}}\left(\mathbf{x}_{t}, t\right).
\end{cases}
\end{equation}

The following loss function is minimized in order to train $\epsilon_\theta$:
\begin{equation}\label{equ:DDPM6}
\min _{\theta} \mathcal{L}(\theta):=\min _{\theta} \mathbb{E}_{\mathbf{x}_{0} \sim q\left(\mathbf{x}_{0}\right), \boldsymbol{\epsilon} \sim \mathcal{N}(\mathbf{0}, \mathbf{I}), t}||\boldsymbol{\epsilon}-\boldsymbol{\epsilon}_{\theta}\left(\mathbf{x}_{t}, t|c\right)||_{2}^{2},
\end{equation}
where $\mathbf{x}_{t}=\sqrt{\alpha_{t}} \mathbf{x}_{0}+\left(1-\alpha_{t}\right) \boldsymbol{\epsilon}$. The noise vector added to its noisy input $x_t$ by the denoising function $\epsilon_\theta$ is estimated, down-weighting the relevance of words at tiny $t$ or low noise levels, and can be thought of as a negative log-likelihood weighted variational restriction.

\section{Method}
\begin{figure}[t]
\centering
\includegraphics[width=0.45\textwidth]{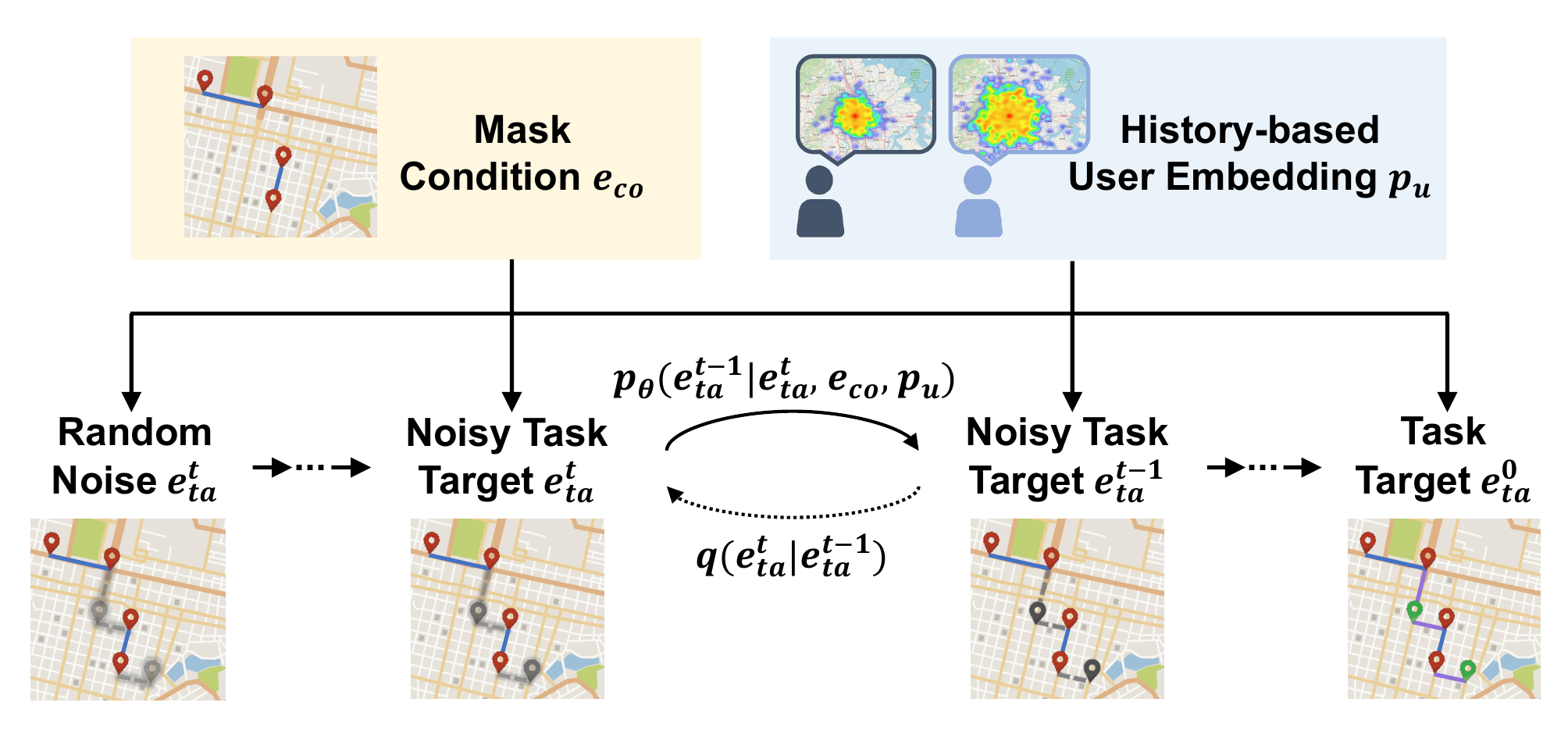}
\caption{The procedure of mobility modeling with GenMove. The reverse process $p_\theta$ gradually converts random noise into trajectory embedding, conditioned on mask condition $e_{co}$ and history-based embedding $p_u$.}
\label{fig:diffusion}
\vspace{-0.5cm}
\end{figure}

\begin{figure*}[t]
\centering
\includegraphics[width=0.95\textwidth]{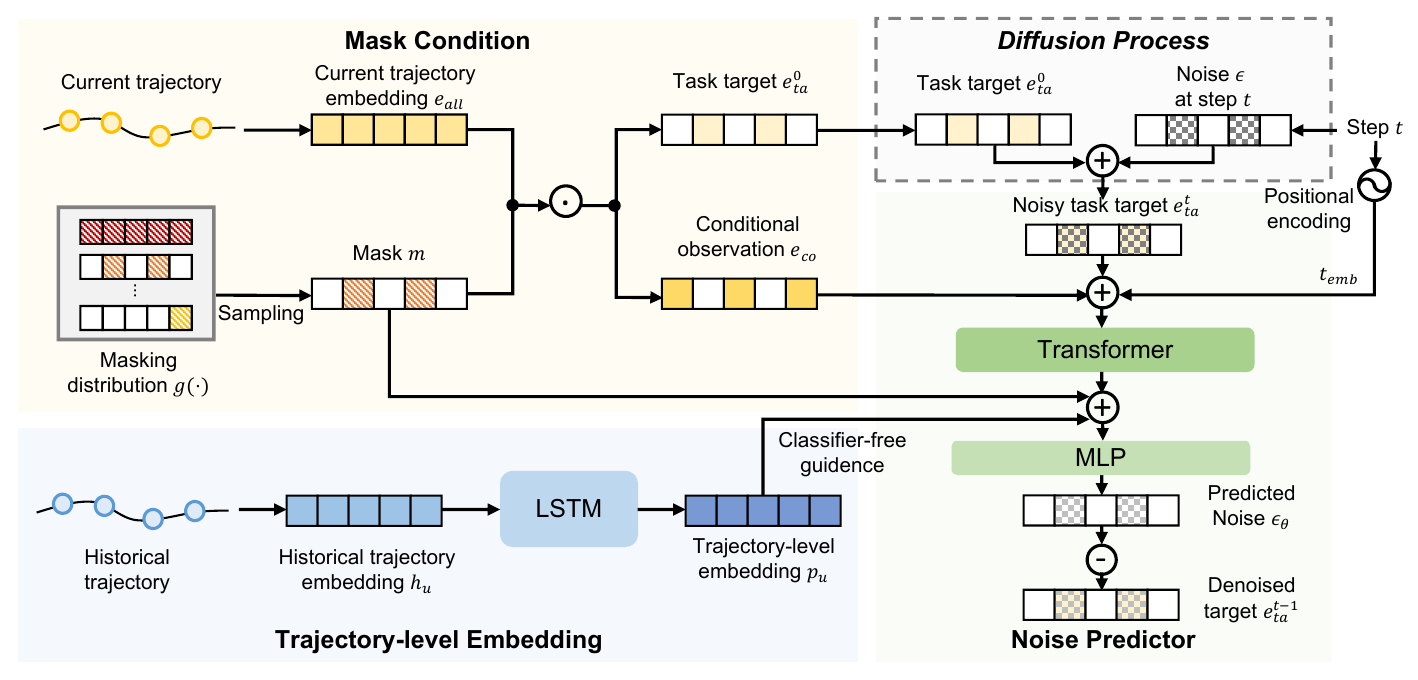}
\caption{The overview architecture of GenMove, which takes the noisy task target $e_{ta}^t$ at the diffusion step $t$, as well as diffusion step embedding $t_{emb}$, the conditional observation $e_{co}$, mask $m$, and user embedding $p_u$ as input. } \label{fig:framework}
\vspace{-0.3cm}
\end{figure*}

Figure~\ref{fig:diffusion} shows the procedure of trajectory modeling with GenMove.
We formulate the learning of the spatiotemporal distribution of trajectories as a denoising diffusion process. In the diffusion process, noise is progressively added to the trajectory until it becomes random noise. During the denoising process, we use $e_{co}$ from the mask condition module and $p_u$ from the contextual trajectory embedding module as conditions to guide the reconstruction of the trajectory from noise. As shown in Figure~\ref{fig:framework}, the overview architecture of GenMove consists of three modules: mask condition, contextual trajectory embedding, and noise predictor. 
To handle diverse formats of different tasks, we utilize masking strategies to unify them with the standardized formats of task targets and conditional observations. To adapt to complex conditions associated with different tasks, we utilize contextual trajectory embedding from historical data to provide essential context for specific tasks. Subsequently, we apply this embedding to noise predictor through a classifier-free guidance approach, allowing the model to adapt its outputs flexibly based on the conditions encountered.



\subsection{Mask Condition}
\label{sec:Mask Condition}
We unify diverse formats of different tasks through masking strategies, including task-specific masks and pattern-general masks.
Specifically, we apply task-specific masks to mask different trajectory parts as task targets for various tasks, enabling a single model to solve multiple mobility modeling tasks: trajectory generation, prediction, and recovery. We utilize pattern-general masks for modeling common mobility patterns across various tasks.

Following the common strategies adopted in previous works~\cite{yuan2023spatio,zhou2023towards}, we perform diffusion and denoising processes on the trajectory embedding. To represent the spatiotemporal dependency of the trajectory, we embed the trajectory into a dense representation as the input for other modules. In particular, to improve the geographical continuity of mobility, we build a spatial graph $G = (V, E)$ that includes a set of nodes $V$ that contains all visited locations, and the weight of each edge represents the Euclidean distance between locations. We utilize the LINE method~\cite{tang2015line} to obtain the current trajectory embedding $e_{all}$.

Then, we construct the mask distribution $g(\cdot)$ by mixing five masking strategies based on adjustable ratios. Each time, we sample from the masking distribution to obtain a specific mask $m$. Therefore, we unify the input forms of all tasks as the current trajectory embedding $e_{all}$ and mask $m$. We transform $e_{all}$ into conditional observation $e_{co}$ and task target $e^0_{ta}$ through the following operations:
\begin{equation}\label{equ:mask}
\begin{cases}
e_{co} = e_{all} \odot m,\\
e^0_{ta} = e_{all} \odot (1-m),
\end{cases}
\end{equation}
where $e_{co}$ denotes conditional observation, $e^0_{ta}$ denotes task target, $m \in \{0,1\}$ denotes mask, and $\odot$ represents element-wise products.

\begin{figure}[h]
\centering
\includegraphics[width=0.8\linewidth]{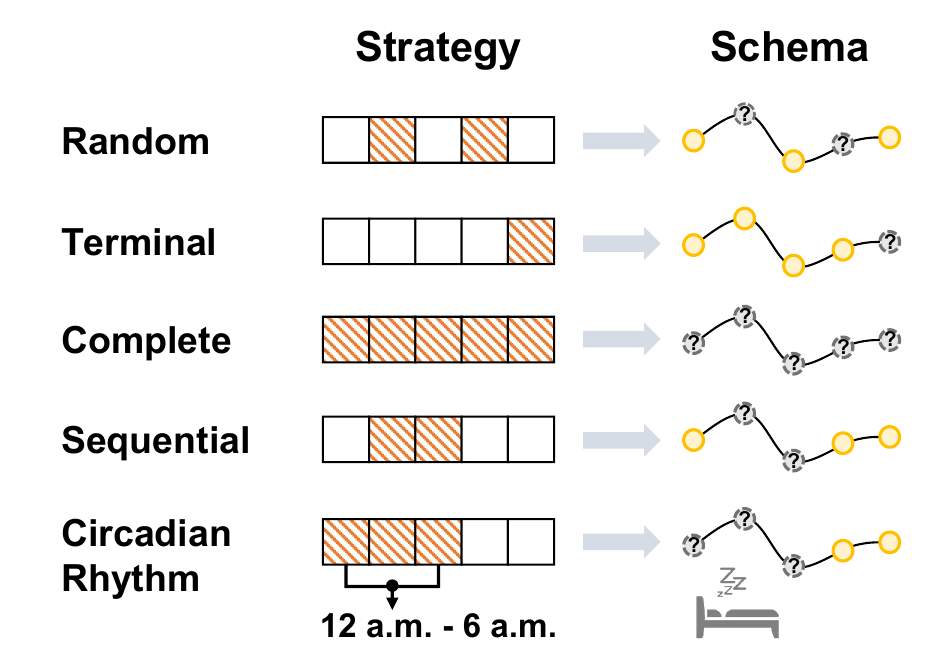}
\caption{Different masking strategies are divided into task-specific masks and mobility-common masks.}
\label{fig:mask}
\vspace{-0.3cm}
\end{figure}

We provide five masking strategies as shown in Figure~\ref{fig:mask}, including task-specific masks corresponding to the features of different tasks, and pattern-general masks for modeling common mobility patterns across various tasks.

(1) Random strategy: This strategy is for the recovery task. It is used when uncertain about missing patterns, selecting recovery targets arbitrarily as a percentage of observed values.

(2) Terminal strategy: This strategy masks future parts of a trajectory, such as the last point or a few points, and is particularly suited for prediction tasks. 

(3) Complete strategy: This strategy is designed for the generation task. This strategy masks the entire trajectory, as no part is needed for observation in generation tasks.

(4) Sequential strategy: This strategy is inspired by the inherent pattern of people's mobility in the real world, where movement is the continuous location transition. It continuously masks a certain percentage of observations.

(5) Circadian rhythm strategy: This strategy aims to capture the circadian rhythm of human mobility, characterized by regular daytime work and nighttime rest patterns. Thus, it masks hours between 12 a.m. and 6 a.m., when people are not frequently moving.

\subsection{Trajectory-level Embedding}
\label{sec:History-based User Embedding}
We design the trajectory-level embedding to provide essential context for different tasks from historical trajectories, adapting to complex conditions associated with different tasks.

We obtain the historical trajectory embedding $h_u$ using the trajectory embedding method. We then use LSTM to map users' historical trajectories to trajectory-level embedding $p_u$, $p_u={\rm LSTM}_{\phi}(h_u)$ enabling users with similar mobility patterns to have similar trajectory-level embedding. The obtained trajectory-level embedding $p_u$ is then introduced into the noise predictor using classifier-free guidance, assisting the trajectory modeling process.

\begin{figure}[t]
\centering
\includegraphics[width=0.4\textwidth]{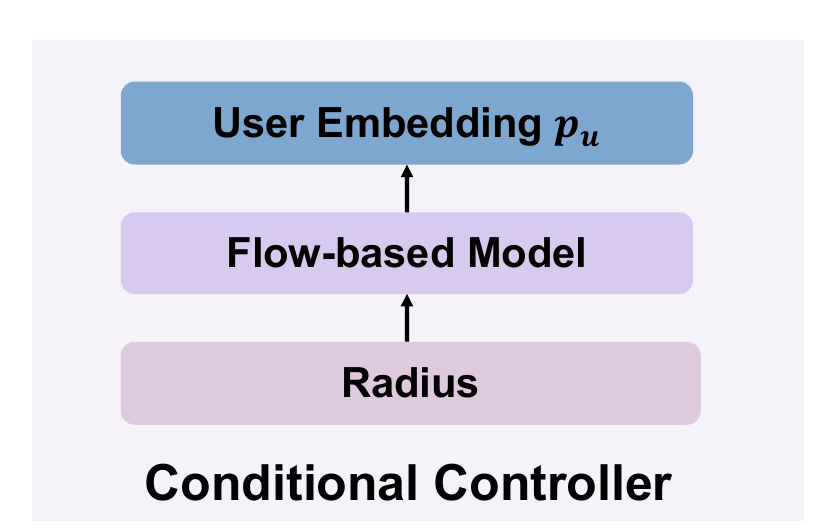}
\caption{Conditional controller replaces History-based user embedding.} 
\label{fig:condition}
\end{figure}

Moreover, we propose \textit{Conditional Controller} to replace \textit{Trajectory-level Embedding}, achieving controllable trajectory generation. Drawing inspiration from GPT 4's ability to integrate plugins for various functions~\cite{liu2023summary}, we design a plugin-like \textit{Conditional Controller} outside the main model, as shown in Figure~\ref{fig:condition}.

Based on the trajectory-level embedding from the \textit{Trajectory-level Embedding} and some features extracted from trajectories (such as radius), we train a conditional controller:
\begin{equation}\label{equ:flow-based model}
p_u={f_{\xi}}(r_u).
\end{equation}
We use the flow-based model ~\cite{zhang2022human} as a generative model $f_\xi(\cdot)$ to produce user embedding $p_u$ based on conditions $r_u$. The $p_u$ output by the \textit{Conditional Controller} serves as a condition for the noise predictor, guiding generation to meet specific conditions, such as a restricted radius of gyration.

\subsection{Noise Predictor}
\label{sec:Denoising Network}
We utilize a transformer to capture the spatio-temporal dependencies of trajectories. Specifically, each step in the denoising process uses the previous prediction value $e^t_{ta}$, the denoising step $t$ with positional encoding, conditional observation $e_{co}$, mask $m$, and user embedding $p_u$ to achieve conditional denoising.

For the diffusion step $t$, we use positional encoding to obtain a 128-dimensional embedding $t_{emb}$ following the previous works~\cite{vaswani2017attention, kong2020diffwave}: 
\begin{equation}\label{equ:diffusion step}
\begin{aligned}
t_{\text {emb}} = & \left[\sin \left(10^{0 \times 4 / 63} t\right), \ldots, \sin \left(10^{63 \times 4 / 63} t\right), \right. \\
& \left.\cos \left(10^{0 \times 4 / 63} t\right), \ldots, \cos \left(10^{63 \times 4 / 63} t\right)\right].
\end{aligned}
\end{equation}
Then, we use the conditional observation $e_{co}$ and the mask $m$ from \textit{Mask Condition} as additional input to the \textit{Noise Predictor}. We introduce the user embedding $p_u$ from \textit{History-based User Embedding} using classifier-free guidance to jointly train conditional and unconditional diffusion models. The \textit{Noise Predictor} takes noisy task target ${e}^t_{ta}$, diffusion step $t$, conditional observation $e_{co}$, mask $m$, and user embedding $p_u$ as input, outputting the predicted noise:
\begin{equation}\label{equ:predicted noise}
\hat{\epsilon}_{\theta}=\epsilon_{\theta}({{e}^t_{ta}}, {e_{co}}, t, m, p_u),
\end{equation}
The output $\hat{\epsilon}_{\theta}$ is of the same shape as ${e}^0_{ta}$ and $\epsilon_{\theta}$. 



\subsection{Training and Sampling}
\label{sec:Training and Sampling}
\paragraph{Training}
We sample different masks from the mask distribution based on a blend of five masking strategies and use these masks to obtain different conditional observation $e_{co}$ mentioned in Section~\ref{sec:Mask Condition}. The conditional observation $e_{co}$ serves as a condition for the conditional diffusion model.
Then, we use classifier-free guidance to train our model. Specifically, we concurrently train an unconditional denoising diffusion model and a conditional denoising diffusion model. A singular neural network is used to parameterize both models. In the case of the unconditional model, a null token $\varnothing$ is used as input for the user embedding $p_u$ during the noise prediction process. We jointly train the unconditional and conditional models by randomly setting $p_u$ to the unconditional class identifier $\varnothing$ with some probability $\lambda$, which is regarded as a hyperparameter in our proposed algorithm.

Based on a similar derivation in Section~\ref{sec:PRELIMINARIES}, we train the model using a simplified loss function:
\begin{equation}\label{equ:diffusion}
\mathcal{L}(\theta) = \mathbb{E}_{{e}^0_{ta},{\epsilon}, t}||\boldsymbol{\epsilon}-\boldsymbol{\epsilon}_{\theta}\left(\mathbf{e}^t_{ta}, t|{e_{co}}, p_u\right)||_{2}^{2}.
\end{equation}

\paragraph{Sampling}
We sample using the following linear combination of the conditional and unconditional model~\cite{ho2022classifier}:
\begin{equation}\label{equ:denoising}
{\tilde{\epsilon}_{\theta}} = (1+\omega){\epsilon}_{\theta}({{e}^t_{ta}}, t|{e_{co}}, p_u) - \omega{\epsilon}_{\theta}({{e}^t_{ta}}, t|{e_{co}}, \varnothing),
\end{equation}
where $\varnothing$ indicates a vector that effectively does not encode any information, with all conditional information set to 0. The model is a diffusion model with just conditional user embedding $p_u$ when $\omega = 0$. The model will prioritize unconditionally predicting noise with a trade-off between variety and sample quality by raising $\omega$. The model is an unconditional diffusion model for $\omega = 1$.

\begin{figure}[t]
\centering
\includegraphics[width=0.4\textwidth]{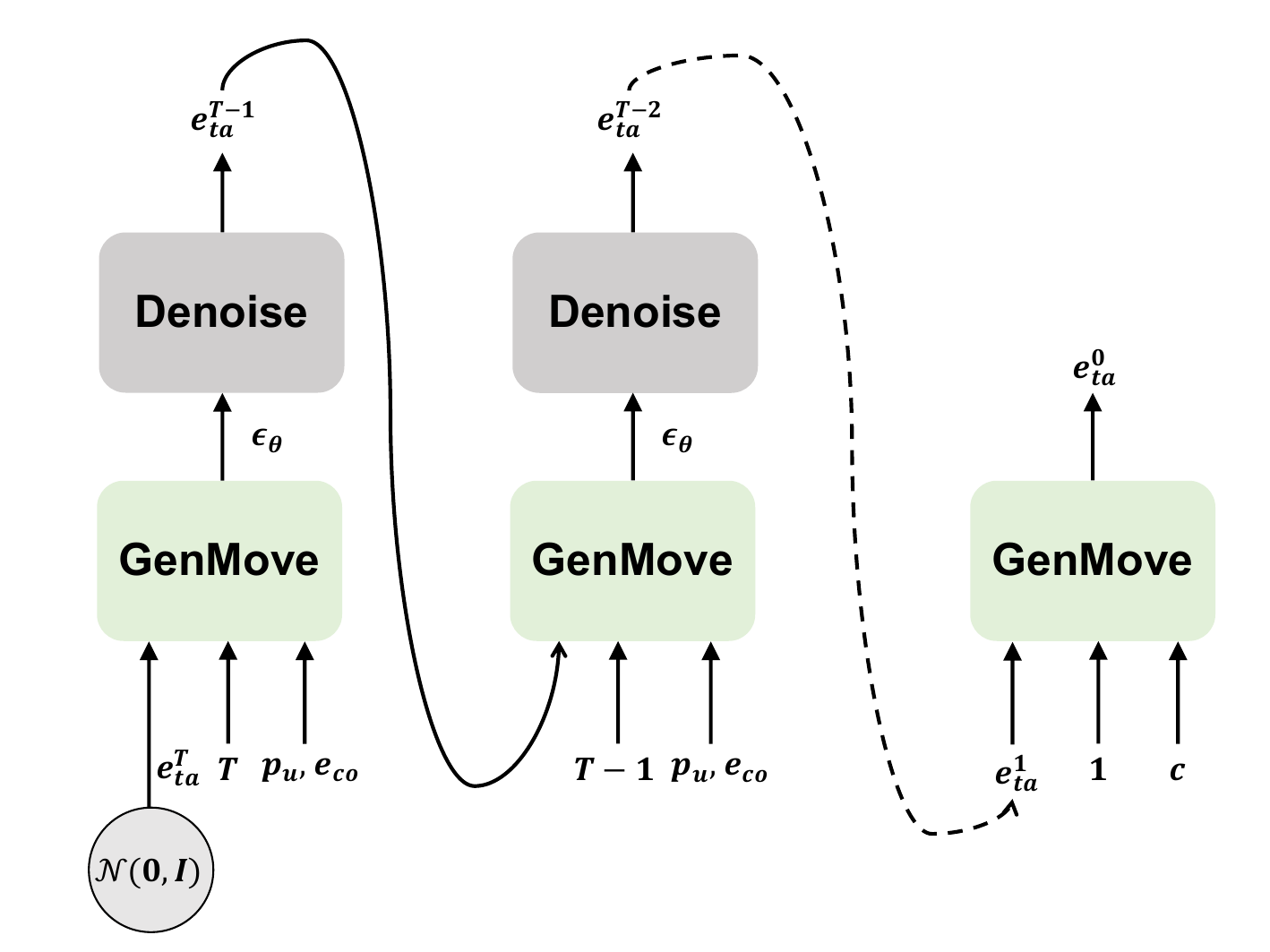}
\caption{The sampling process of GenMove.} 
\label{fig:sample}
\end{figure}

Figure~\ref{fig:sample} illustrates the iterative denoising steps during the sampling process. In this process, the noise predicted at each step is removed, progressively decreasing from $T$ steps to 1 step, ultimately resulting in clean data, denoted as $e^{0}_{ta}$. Below, we list the formula for noise removal at each step:
\begin{equation}\label{equ:sampling}
e^{t-1}_{ta}={\frac{1}{\sqrt{\alpha_{t}}}(e^{t}_{ta}-\frac{1-\alpha_{t}}{\sqrt{1-\bar{\alpha_{t}}}}\tilde{\epsilon}_{\theta})}+{\sigma_t}z
\end{equation}
where $\tilde{\epsilon}_{\theta}$ is a function approximator intended to predict $\epsilon$ from $e^{t}_{ta}$ and $z \sim N (0, I)$.

\section{Experiments}

\subsection{Experimental Settings}
\subsubsection{Dataset}
We conducted broad experiments on two real-world datasets (\textbf{ISP} and \textbf{MME}), which were collected from two cities in China (Shanghai and Nanchang).
The details of datasets are summarized in Table~\ref{table:datasets}.
\begin{itemize}[leftmargin=*]
\item \textbf{ISP}: The ISP dataset is sourced from a prominent Internet service provider in China. The dataset contains trajectories of over 90000 users in Shanghai over a week, each containing an anonymous user ID, timestamp, and cellular base station.
\item \textbf{MME}: The MME dataset is provided by an operator in China. The dataset contains the trajectories of more than 6000 users over a week in Nanchang, with each mobility record containing an anonymous user ID, time stamp, and region ID.
\end{itemize}

Moreover, we have taken proactive measures to protect users' privacy in our dataset. Firstly, all data collected in this study are stored on secure data warehouse servers behind the company's firewall. Secondly, staff from our partners, ISP and MME, have anonymized all user information, including personal and location identifiers.

\begin{table}[t]
\centering
\caption{Basic statistics of mobility datasets.}
\label{table:datasets}
\scalebox{0.9}{
\begin{tabular}{ >{\centering\arraybackslash}m{1cm} 
>{\centering\arraybackslash}m{1cm} 
>{\centering\arraybackslash}m{1.2cm} 
>{\centering\arraybackslash}m{1cm}
>{\centering\arraybackslash}m{1cm}
>{\centering\arraybackslash}m{1cm}}
\toprule
Dataset & City & Duration & \#Users & \#Loc & \#Traj\\ 
\midrule
ISP & Shanghai & 7 days & 90037 & 9158 & 261042\\ 
MME & Nanchang & 7 days & 6218 & 4096 & 43967\\ 
\bottomrule
\end{tabular}
}
\end{table}

\subsubsection{Tasks}
\begin{figure}[t]
\centering
\includegraphics[width=0.95\linewidth]{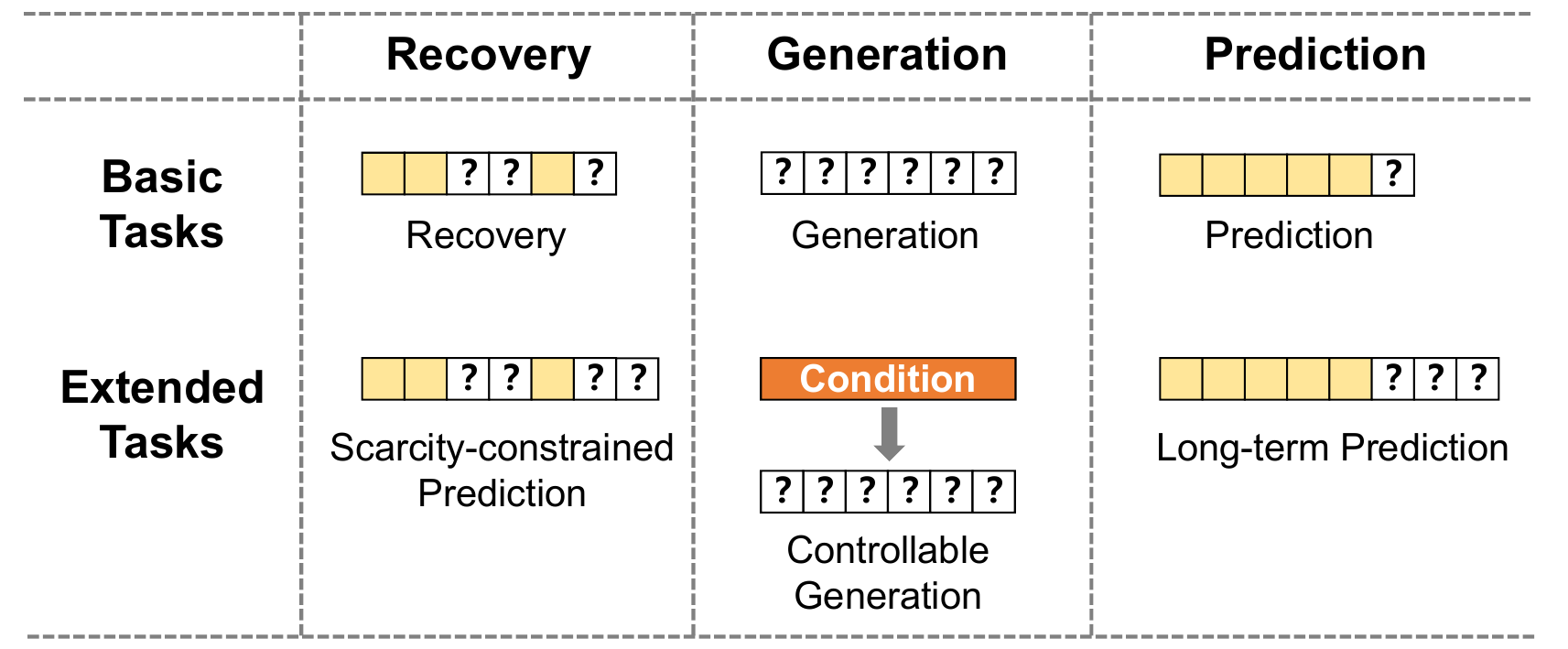}
\caption{Six trajectory-related tasks, divided into basic and extended tasks.}
\label{fig:task}
\end{figure}

As shown in Figure~\ref{fig:task}, we evaluate our method across six typical trajectory-related tasks. However, our model is not limited to these tasks; it can adapt to a broader range of tasks. These six tasks are categorized into generation, recovery, and prediction. Each type is further divided into basic and extended tasks. For example, in the generation category, the basic task is the unconditional generation, while the extended task is the controllable generation (i.e., generating trajectories that meet specific conditions). In the prediction category, the basic task involves the next location prediction, while the extended task involves long-term prediction (predicting locations over multiple time steps). In the recovery category, the basic task is trajectory recovery, and the extended task is scarcity-constrained prediction (i.e., the next location prediction based on sparse trajectories).

\subsubsection{Metrics}
Following the previous related works~\cite{feng2020learning,wang2023pategail,xia2021attnmove,sun2021periodicmove,feng2018deepmove,luo2021stan}, we select common metrics for mobility modeling tasks to evaluate the effectiveness of our model. 

\paragraph{Trajectory Generation} We evaluate the generated trajectory using the following six metrics:
\begin{itemize}[leftmargin=*]
\item \textbf{Distance}: This is travel distance, which is determined by adding up each user's trip distance over a certain period.
\item \textbf{Radius}: The spatial range of the user's daily movement is represented by the radius of gyration.
\item \textbf{Duration}: This refers to the duration of stay, measured as the time spent at each visited destination.
\item \textbf{Daily-loc}: This refers to the daily frequency of visited locations, quantified as the number of locations each user trips every day.
\item \textbf{Density}: This metric evaluates the geographical distribution by comparing the density of the total and actual trajectories on a global scale.
\item \textbf{Trip}: At the trajectory level, this statistic measures the relationship between the starting and ending points of a trip. It involves calculating the probability distributions for the initial and final locations in both the original and generated trajectories.
\end{itemize}
We employ the Jensen-Shannon divergence (JSD) to visually assess the similarity between the generated sequences and the real sequences. Specifically, for two distributions, $\boldsymbol{m}$ and $\boldsymbol{n}$, the Jensen-Shannon Divergence (JSD) between them is defined as:
\begin{equation}\label{equ:JSD}
{\rm JSD}(\boldsymbol{m},\boldsymbol{n})=\frac{1}{2}{\rm KL}(\boldsymbol{m}||\frac{\boldsymbol{m}+\boldsymbol{n}}{2})+\frac{1}{2}{\rm KL}(\boldsymbol{n}||\frac{\boldsymbol{m}+\boldsymbol{n}}{2}),
\end{equation}
where ${\rm KL}(\cdot||\cdot)$ is the Kullback-Leibler divergence~\cite{torabi2018behavioral}.

\paragraph{Trajectory Recovery} We evaluate the recovered trajectory using the following three metrics:
\begin{itemize}[leftmargin=*]
\item \textbf{Recall}: It evaluates the model's ability to recover all ground-truth locations, averaged over all tested values. If the Recall is 1, it means that all ground-truth locations are fully recovered, as opposed to 0 if they are not recovered at all.
\item \textbf{MAP}: It assesses the global ranking task to determine the quality of the entire ranking list, encompassing all locations. A higher MAP number indicates superior performance.
\item \textbf{Distance}: It is defined as the geographical distance between the recovery location's center and the ground truth. The performance will improve as the distance decreases.
\end{itemize}

\paragraph{Trajectory Prediction}
We established Accuracy@k as our assessment metric. To determine the top-k prediction accuracy, we initially employ the model to produce the k most probable predictions derived from the provided historical data, subsequently verifying if the actual subsequent location is among these top-k projections.

We repeated the experiments for each task five times, and the performance metrics presented are the average values of these five experiments.

\subsubsection{Baselines}
We compare the performance of our model with state-of-the-art baselines for different basic tasks. The baseline for the extended task is derived from a combination of one or more best baselines capable of completing the task.
\paragraph{Trajectory Generation} The baselines for trajectory generation are as follows:
\begin{itemize}[leftmargin=*]
\item \textbf{TimeGEO}~\cite{jiang2016timegeo} is a model-based trajectory synthesis method that characterizes temporal choices through dwell rate, burst rate, and the weekly home-based tour frequency, while employing the explore and preferential return (EPR) model~\cite{song2010modelling} to represent spatial choices.
\item \textbf{MoveSim}~\cite{feng2020learning}: This generative adversarial approach integrates the domain knowledge of human movement patterns.
\item \textbf{VOLUNTEER}~\cite{long2023practical}: This model consists of a two-layer structure of user VAE and trajectory VAE to model the joint distribution of user attributes and mobility behavior to generate more practical trajectories.
\item \textbf{PateGail}~\cite{wang2023pategail}: It employs the robust generative adversarial imitation learning model to replicate human decision-making processes.
\item \textbf{DiffTraj}~\cite{zhu2023diffusion}: It utilizes diffusion models to generate trajectories by using the spatiotemporal features of real trajectories as conditional information
\end{itemize}

\paragraph{Trajectory Recovery} The baselines for trajectory recovery are as follows:
\begin{itemize}[leftmargin=*]
\item \textbf{Linear}~\cite{hoteit2014estimating}: This approach presumes that a linear function can estimate the missing trajectory points. It is a technique based on rules with a significant assumption.
\item \textbf{TrImpute}~\cite{elshrif2022network}: This approach depends on collective intelligence, using adjacent GPS coordinates to facilitate recovery.
\item \textbf{RF}~\cite{li2019reconstruction}: The random forest utilizes all elements of each trajectory for recovery. The random forest classifier is developed by training on features such as the entropy and radius of each trajectory, the gaps in time where data is missing, and the locations immediately before and after these gaps.
\item \textbf{AttnMove}~\cite{xia2021attnmove}: It employs diverse intra-trajectory and inter-trajectory attention processes to enhance the modeling of user movement patterns and capitalizes on periodic patterns in long-term history to reconstruct trajectories.
\item \textbf{PeriodicMove}~\cite{sun2021periodicmove}: This is an innovative neural attention model based on a graph neural network that systematically captures intricate location transition patterns, multilevel periodicity, and the evolving periodicity of human movement.
\end{itemize}

\paragraph{Trajectory Prediction} The baselines for trajectory prediction are as follows:
\begin{itemize}[leftmargin=*]
\item \textbf{Markov Model}~\cite{gambs2012next}: The Markov model is a statistical framework used to characterize state transitions across time. It employs previous trajectory data for position forecasting by computing the transition probabilities among different locations.
\item \textbf{DeepMove}~\cite{feng2018deepmove}: The method develops a multimodal embedding recurrent neural network to encapsulate intricate sequential transitions by concurrently embedding several elements that influence human mobility.
\item \textbf{STAN}~\cite{luo2021stan}: This approach correlates non-adjacent but functionally analogous visited locations to forecast the next location.
\item \textbf{DSTPP}~\cite{yuan2023spatio}: The approach employs diffusion models to comprehend intricate spatio-temporal joint distributions for trajectory prediction.

\end{itemize}

\paragraph{Task-general Model} The baselines are a general model capable of accomplishing three tasks: trajectory generation, recovery, and recovery.
\begin{itemize}[leftmargin=*]
\item \textbf{LSTM}~\cite{Kong2018HST}:  The LSTM network excels in processing sequential data and can encode long-term dependencies, which can naturally be applied to different tasks.
\item \textbf{TrajGDM}~\cite{chu2024simulating}: This method is a trajectory generation framework using the diffusion model to encapsulate the universal mobility pattern inside a trajectory dataset.
\end{itemize}

\begin{table*}[t]
\small
\centering
\caption{Performance comparisons of the trajectory generation task on two mobility datasets.}
\scalebox{0.89}{
\begin{tabular}{lcccccccccccc}
\toprule
& \multicolumn{6}{c}{\textbf{ISP}} & \multicolumn{6}{c}{\textbf{MME}} \\
\cmidrule(lr){2-7} \cmidrule(lr){8-13}
& \textbf{Distance} & \textbf{Radius} & \textbf{Duration} & \textbf{Daily-loc}  & \textbf{Density} & \textbf{Trip}  
& \textbf{Distance} & \textbf{Radius} & \textbf{Duration} & \textbf{Daily-loc}& \textbf{Density} & \textbf{Trip}  \\
\midrule
TimeGEO & 0.0628 & 0.4703 & 0.0912 & 0.4257 &  0.0295 & 0.0726 & 0.0443 & 0.4126 & 0.0623 & 0.3868 &  0.0320 &  0.0619\\
LSTM  & 0.0712&  0.0540&  0.1185&  0.4051& 0.0258 & 0.0682 &0.0362&  0.3987&  0.0601&  0.3247 & 0.0309 & 0.5327\\
MoveSim & 0.0254 & 0.4021 & 0.0635 & 0.2802& 0.0247 &  0.0658 & 0.0217 & 0.3572 & 0.0561 & 0.2526 & 0.0294 & 0.0451\\
VOLUNTEER & 0.0137 & 0.4316 & 0.0453 & 0.1584 & 0.0217 & 0.0506 & 0.0098 & 0.3725 & 0.0507 & 0.1435 & 0.0215 & 0.0403\\
PateGail & 0.0102 & 0.2125 & 0.0384 & 0.0630 & 0.0203 & 0.0358 & 0.0004 & 0.1806 & 0.0783 & 0.0887 & 0.0168 & 0.0310\\
DiffTraj & \underline{0.0085} & \underline{0.1736} & \underline{0.0307} & \underline{0.0541} & 0.0158  & 0.0280
         & \underline{0.0002} & \underline{0.1657} & \underline{0.0498} & \underline{0.0653} & 0.0105 & 0.0241\\
TrajGDM  & 0.0089	& 0.1864  & 0.0326  & 0.0571 & \underline{0.0139} & \underline{0.0224}  & 0.0003 & 0.1701 & 0.0515 & 0.0712 & \underline{0.0092} & \underline{0.0254}\\
GenMove & \textbf{0.0074} & \textbf{0.1431} & \textbf{0.0262} & \textbf{0.0488} & \textbf{0.0122} & \textbf{0.0249}
     & \textbf{0.0002} & \textbf{0.1512} & \textbf{0.0467} & \textbf{0.0507}& \textbf{0.0078} & \textbf{0.0181} \\
\bottomrule
\end{tabular}
}
\label{tab:Trajectory Generation comparison}
\end{table*}

\begin{table*}[t]
\centering
\begin{minipage}{0.49\textwidth}
\caption{Performance comparisons of the trajectory prediction task on two mobility datasets.}
\label{tab:Next Location Prediction comparison}
\resizebox{\textwidth}{!}{ 
\begin{tabular}{lcccccc}
\toprule
& \multicolumn{3}{c}{\textbf{ISP}} & \multicolumn{3}{c}{\textbf{MME}} \\ 
\cmidrule(lr){2-4} \cmidrule(lr){5-7}
& \textbf{Acc@1} & \textbf{@3} & \textbf{@5} & \textbf{Acc@1} & \textbf{@3} & \textbf{@5} \\
\midrule
Markov & 0.3164 & 0.4215 & 0.5323 & 0.3832 & 0.4768 & 0.6027 \\
LSTM & 0.3772 & 0.4581 & 0.5907 & 0.4325 & 0.5152 & 0.6590 \\
DeepMove & 0.4013 & 0.4872 & 0.6391 & 0.4683 & 0.5324 & 0.6838 \\
STAN & 0.4175 & 0.4946 & 0.6503 & 0.4819 & 0.5506 & 0.7014 \\
DSTPP & \underline{0.4201} & \underline{0.4968} & \underline{0.6510} & \underline{0.4838} & \underline{0.5520} & \underline{0.7022} \\
TrajGDM  & 0.4182 & 0.4950	& 0.6501 & 0.4768 & 0.5430	& 0.6917\\
GenMove & \textbf{0.4248} & \textbf{0.5015} & \textbf{0.6596} & \textbf{0.4865} & \textbf{0.5552} & \textbf{0.7068} \\
\bottomrule
\end{tabular}
}
\end{minipage}
\hfill
\begin{minipage}{0.49\textwidth}
\caption{Performance comparisons of the trajectory recovery task on two mobility datasets.}
\label{tab:Trajectory Recovery comparison}
\resizebox{\textwidth}{!}{ 
\begin{tabular}{lcccccc}
\toprule
& \multicolumn{3}{c}{\textbf{ISP}} & \multicolumn{3}{c}{\textbf{MME}} \\ 
\cmidrule(lr){2-4} \cmidrule(lr){5-7}
& \textbf{Recall} & \textbf{MAP} & \textbf{Dis.} 
& \textbf{Recall} & \textbf{MAP} & \textbf{Dis.} \\
\midrule
Linear & 0.5024 & 0.5385 & 1709 & 0.5259 & 0.5649 & 1364 \\
TrImpute & 0.5406 & 0.5963 & 1602 & 0.5723 & 0.6216 & 1219 \\
RF & 0.4025 & 0.4397 & 2215 & 0.4371 & 0.4686 & 1752 \\
LSTM & 0.5210 & 0.6153 & 1583 & 0.6015& 0.6741 & 1205 \\
AttnMove & 0.5893 & 0.6612 & 1436 & 0.6427 & 0.7153 & 1086 \\
PeriodicMove & \underline{0.5982} & \underline{0.6748} & \underline{1372} & \underline{0.6548} & \underline{0.7225} & \underline{993} \\
TrajGDM  &0.5715   & 0.6454	 & 1597 & 0.6181 & 0.6678 & 1125\\
GenMove & \textbf{0.6041} & \textbf{0.6856} & \textbf{1304} & \textbf{0.6674} & \textbf{0.7302} & \textbf{925} \\
\bottomrule
\end{tabular}
}
\end{minipage}
\end{table*}

\subsection{Hyperparameters Settings}
The dataset is divided by the user, with the first 70\% as the training set, the second 10\% as the validation set, and the remaining 20\% as the test set. As for the hyperparameter settings of our model, we set the embedding size of the diffusion step and the size
of the location embedding to 128, the number of channels of the convolutional network to 64, the number of transformer layers to 4, and the number of heads of the attention mechanism to 8. Then, we set the learning rate to 1e-3, the diffusion steps to 1000, and the batch size to 16 during the training process. The details of hyperparameters settings are shown in Table~\ref{tab::hyper-parameters}.
Finally, the proposed framework is implemented with Pytorch. We train it on a Linux server with eight GPUs (NVIDIA RTX 2080 Ti * 8). 
\begin{table}[t]
	\centering
	\caption{The details of hyperparameters settings.}
	\setlength\tabcolsep{4pt}
	\scalebox{1.0}{
	\begin{tabular}{l|c}
		\toprule[1pt]
		\textbf{Hyper-parameters} & \textbf{Settings}\\ \hline
		The embedding size of diffusion step & 128 \\ \hline
		The embedding size of location embedding & 128 \\ \hline
            The channel number of the convolutional network & 64 \\ \hline
            Transformer layer number & 4 \\ \hline
            Transformer head number & 8 \\ \hline
            Learning rate & 1e-3 \\ \hline
            Diffusion steps & 1000 \\ \hline
	    Batch size & 16\\ 
	\bottomrule[1pt]
	\end{tabular}}
	\label{tab::hyper-parameters}
	\vspace{-0cm}
\end{table}

\subsection{Overall Performance}
\begin{figure}[t]
\centering
\includegraphics[width=0.9\linewidth]{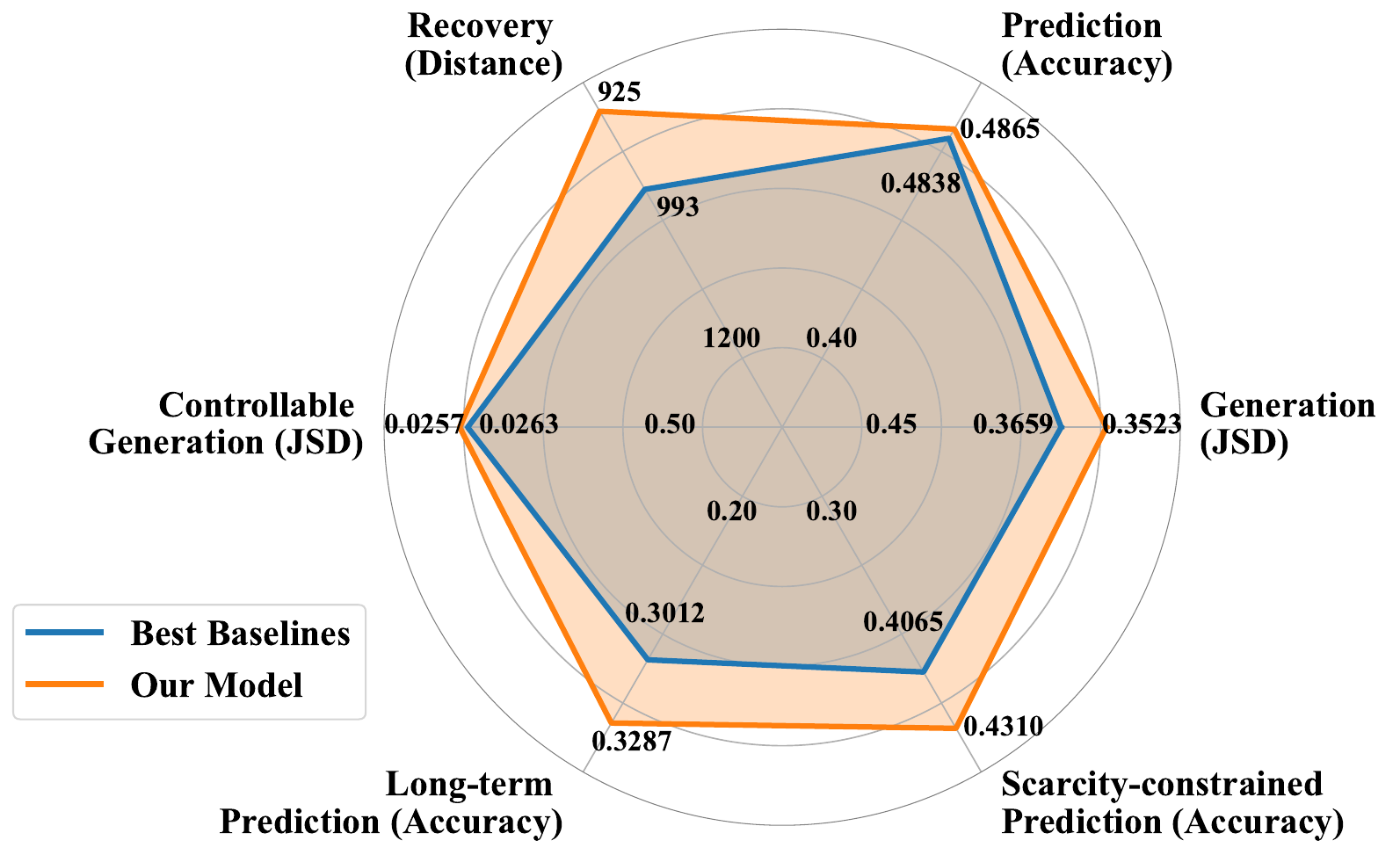}
\caption{Model performance comparison on MME dataset. } 
\label{fig:MME performance}
\end{figure}

As shown in Figure~\ref{fig:Performance} and~\ref{fig:MME performance}, our method demonstrates similar or better performance than the state-of-the-art baselines for all tasks on two datasets. This reflects that our general framework is capable of performing various tasks with optimal performance.

\subsubsection{Trajectory Generation}
\begin{figure}[t]
\centering
\subfloat[Real trajectories]{\includegraphics[width=.20\textwidth]{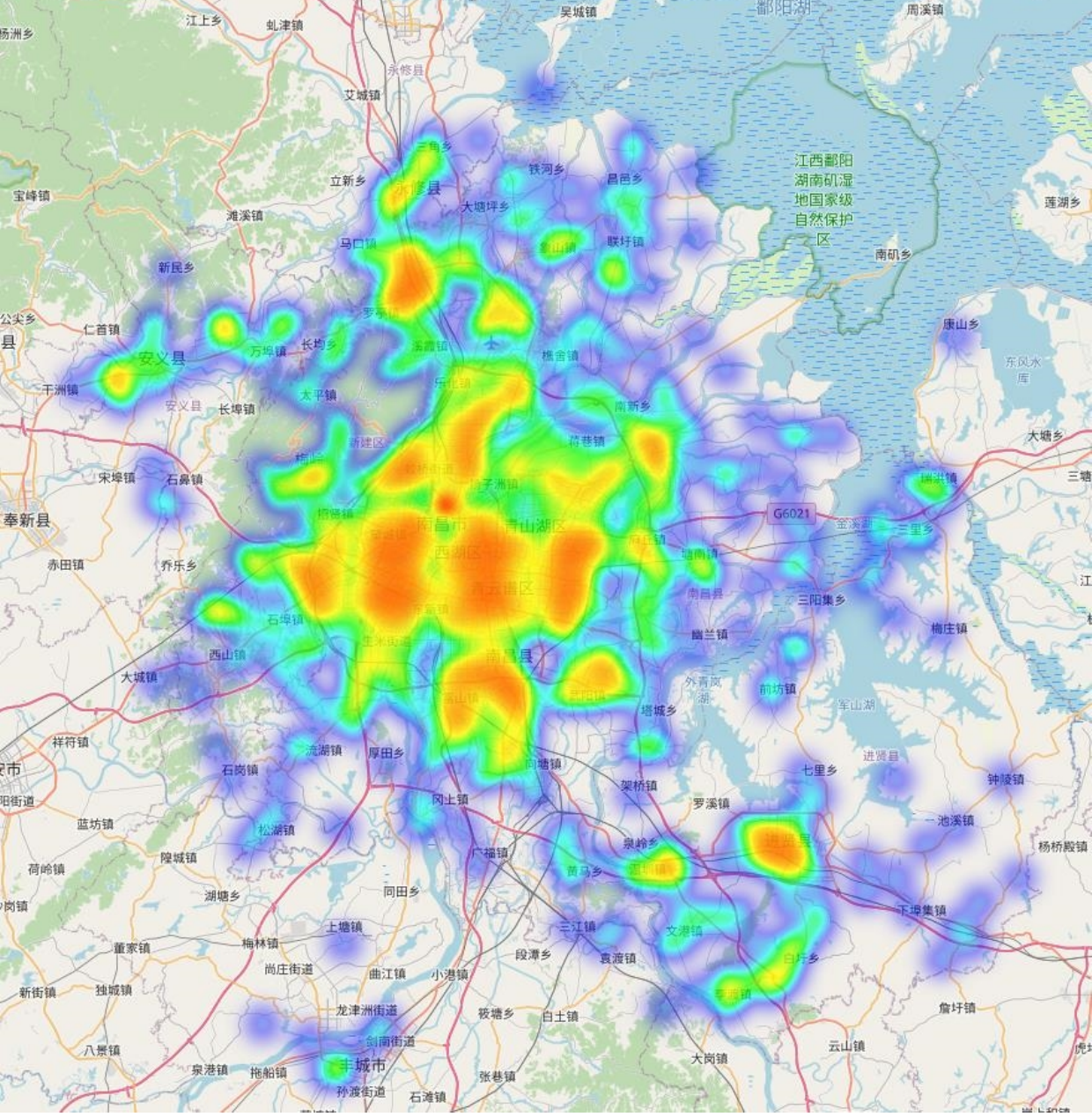}}
\hfil
\subfloat[Generated trajectories based on GenMove]{\includegraphics[width=.20\textwidth]{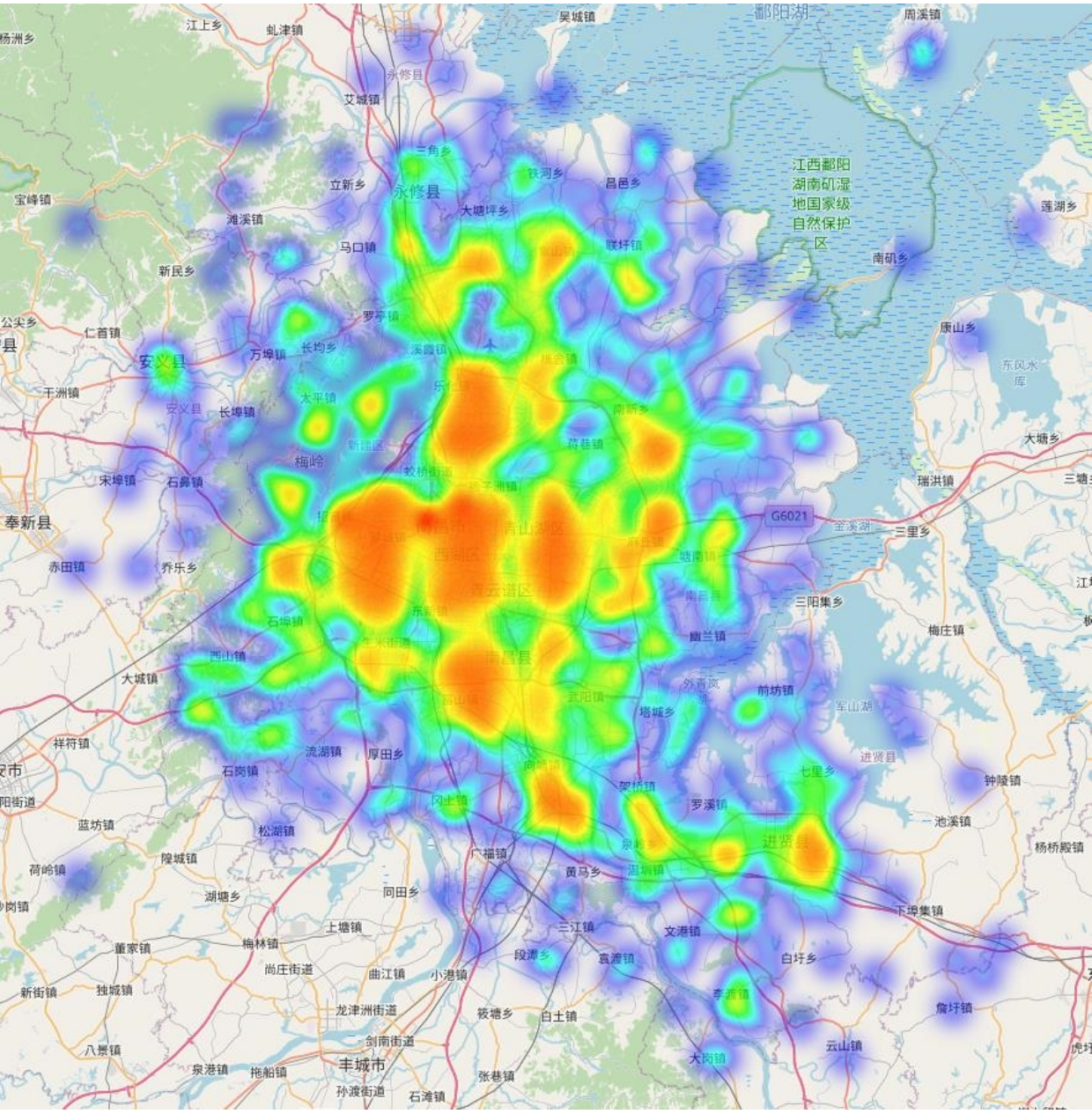}}
\hfil
\subfloat[Generated trajectories based on DiffTraj]{\includegraphics[width=.20\textwidth]{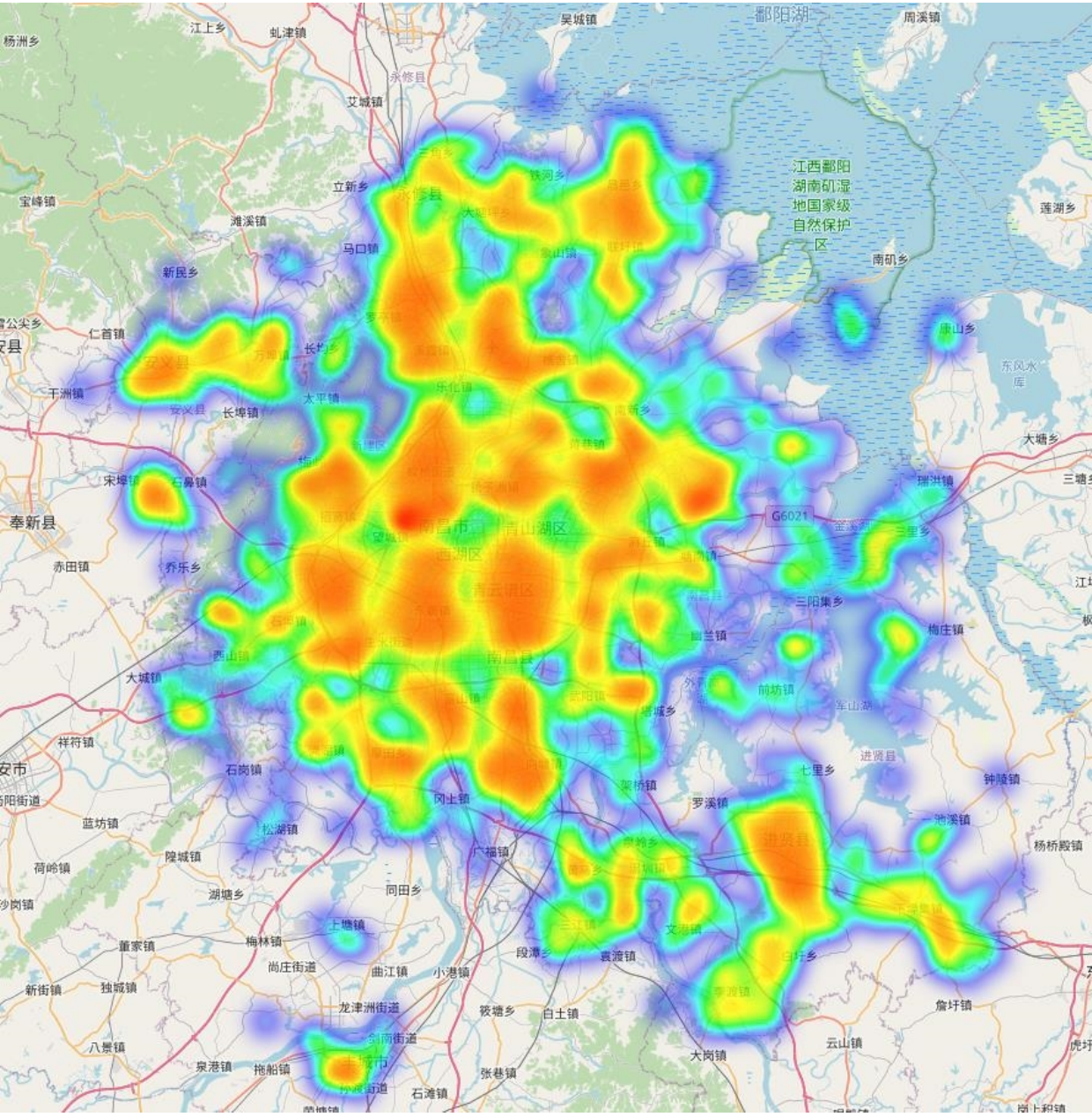}}
\hfil
\subfloat [Generated trajectories based on PateGail]{\includegraphics[width=.20\textwidth]{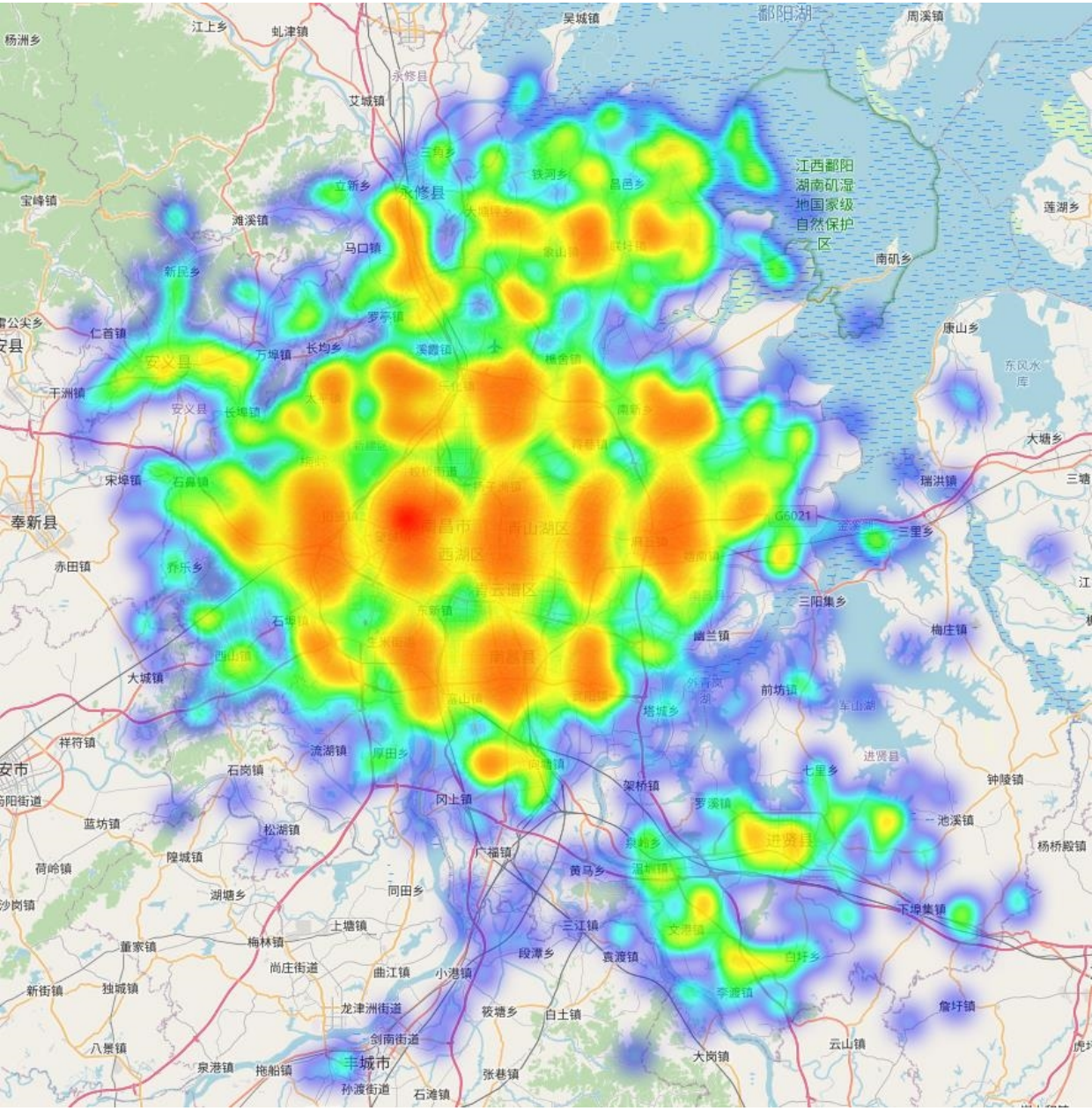}}
\caption{Comparison of heatmaps between real and generated trajectories on the MME dataset.}
\label{fig:visualization}
\end{figure}

\begin{figure}[t]
\centering
\subfloat [Radius of 15 kilometers]{\includegraphics[width=.20\textwidth]{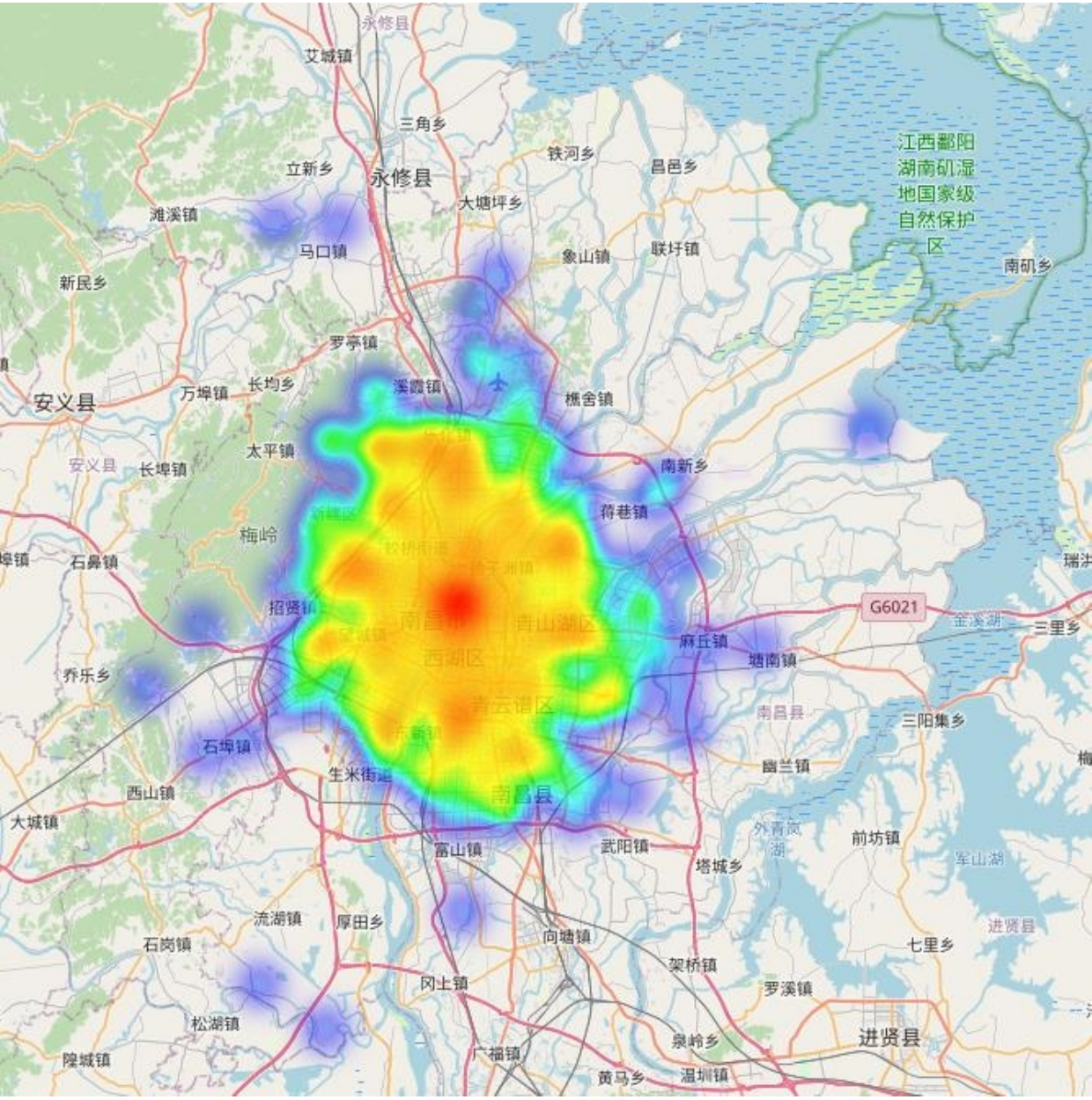}}
\hfil
\subfloat [Radius of 45 kilometers]{\includegraphics[width=.20\textwidth]{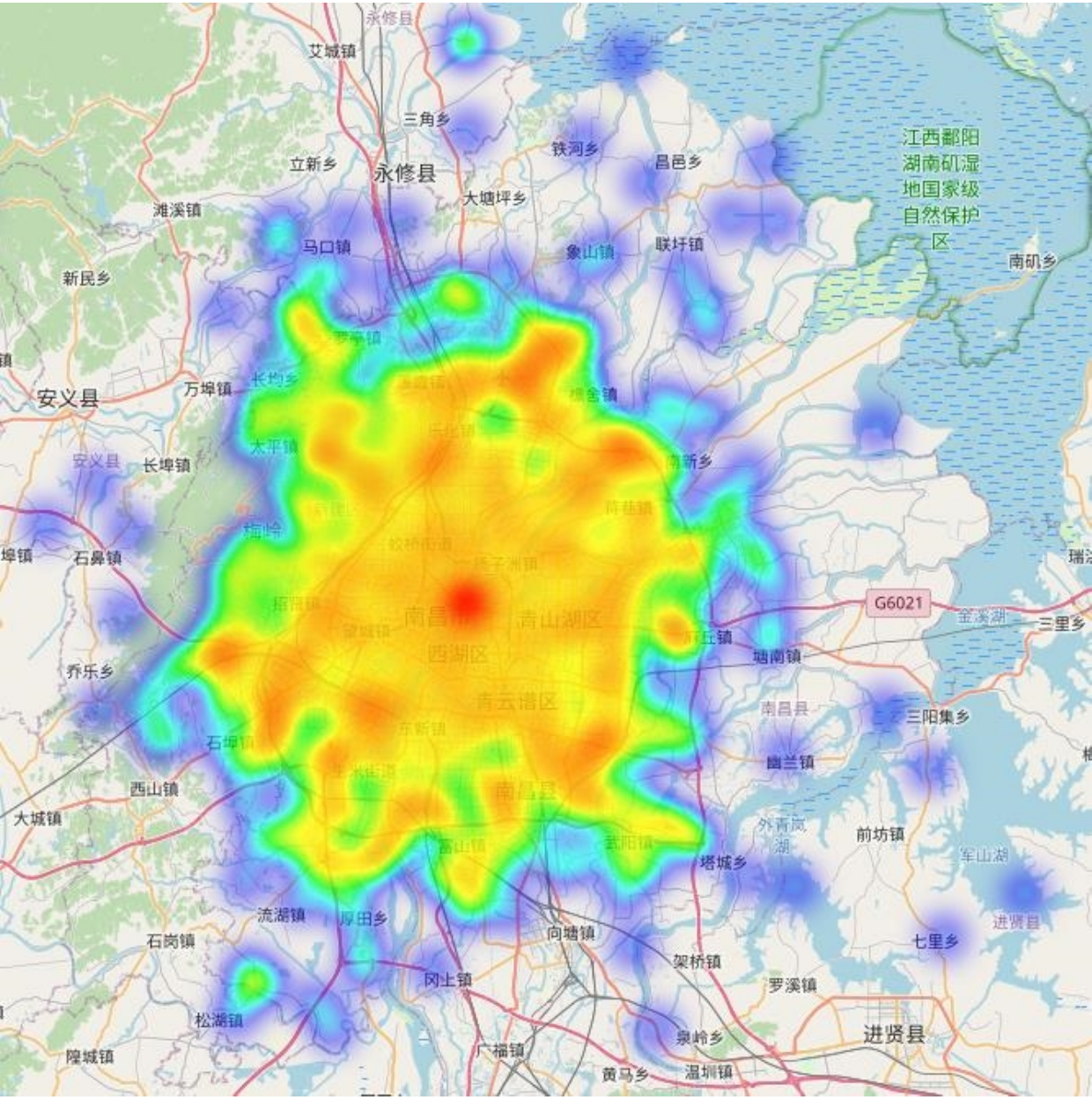}}
\caption{Heatmap of controllable trajectory generation conditioned on the radius of gyration.}
\label{fig:Conditional-Trajectory-Generation}
\end{figure}

Table~\ref{tab:Trajectory Generation comparison} shows the performance in the basic task of unconditional generation, where our method outperformed others across all metrics, with an average performance difference of about 13\% compared to various baselines. 
We also visualized the generated results to better present the performance, as shown in Figure~\ref{fig:visualization}. Our model's generated trajectory distribution resembles the real trajectory distribution the most. The distribution generated by the baseline is more uniform and fails to replicate the trajectory distribution accurately. This reflects that the trajectories generated by our model can more accurately simulate real mobility patterns.
Considering the demand for controllable generation in practical applications, we further conducted experiments on controllable generation. Figure~\ref{fig:Conditional-Trajectory-Generation} shows the effect of our controllable generation. Specifically, we use different radius constraints to control the generation. We find that setting different radius for controllable generation results in trajectories with distinctly different ranges.

\begin{figure}[t]
\centering
\subfloat[Scarcity-constrained prediction on MME dataset]{\includegraphics[width=.23\textwidth]{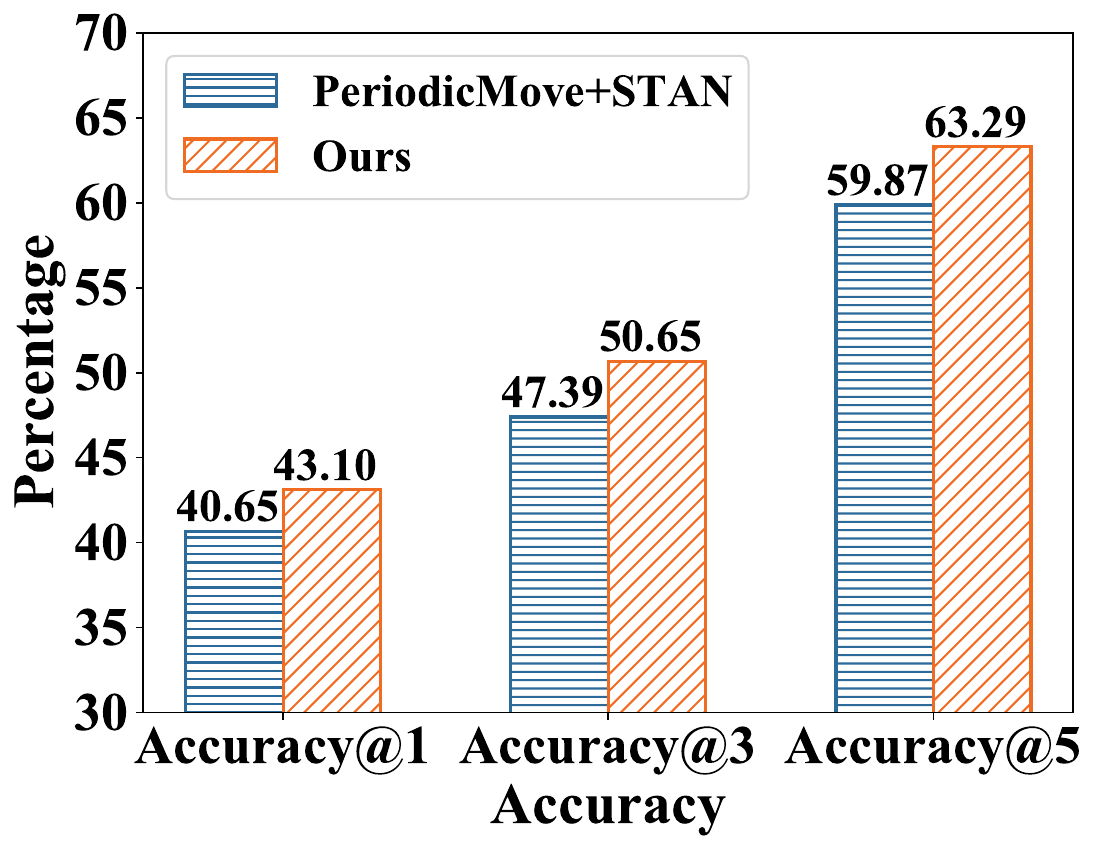}}
\hfil
\subfloat[Scarcity-constrained prediction on ISP dataset]{\includegraphics[width=.23\textwidth]{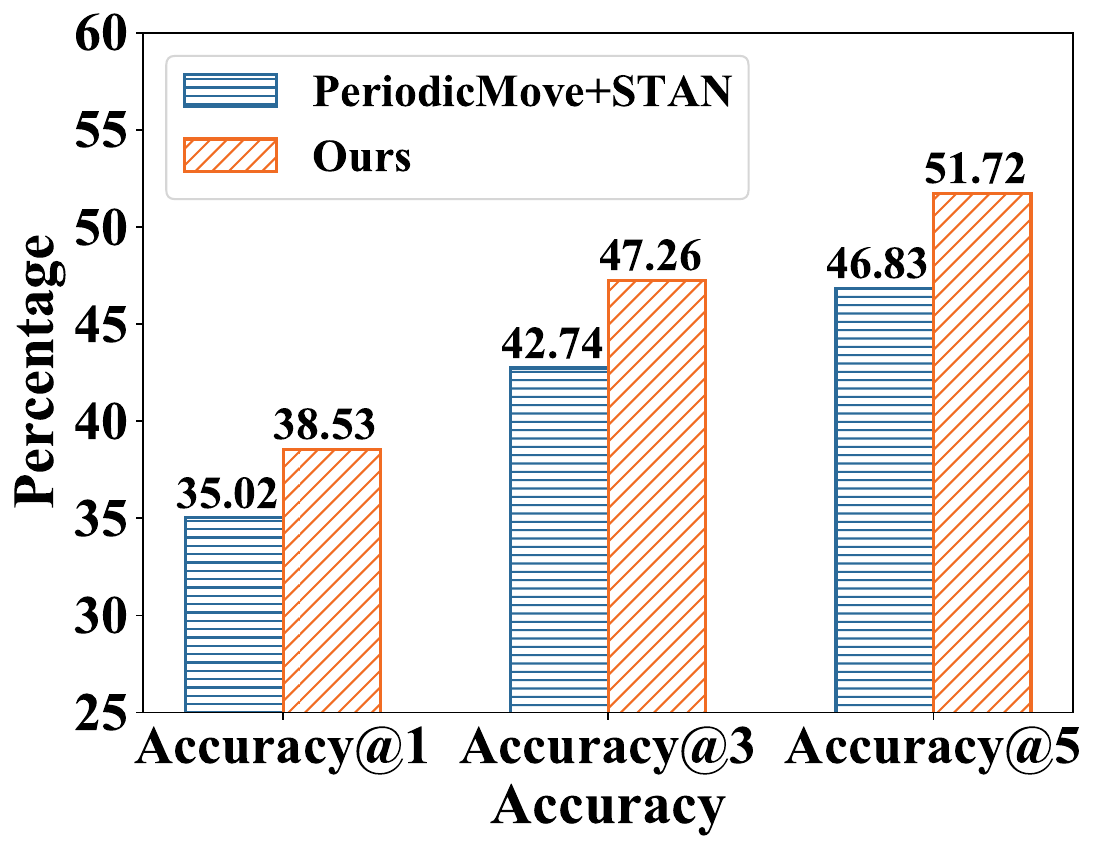}}
\caption{Performance of scarcity-constrained trajectory prediction on two datasets.} 
\label{fig:incomplete-prediction}
\end{figure}

\begin{figure}[h]
\centering
\subfloat[Long-term prediction on MME dataset]{\includegraphics[width=.23\textwidth]{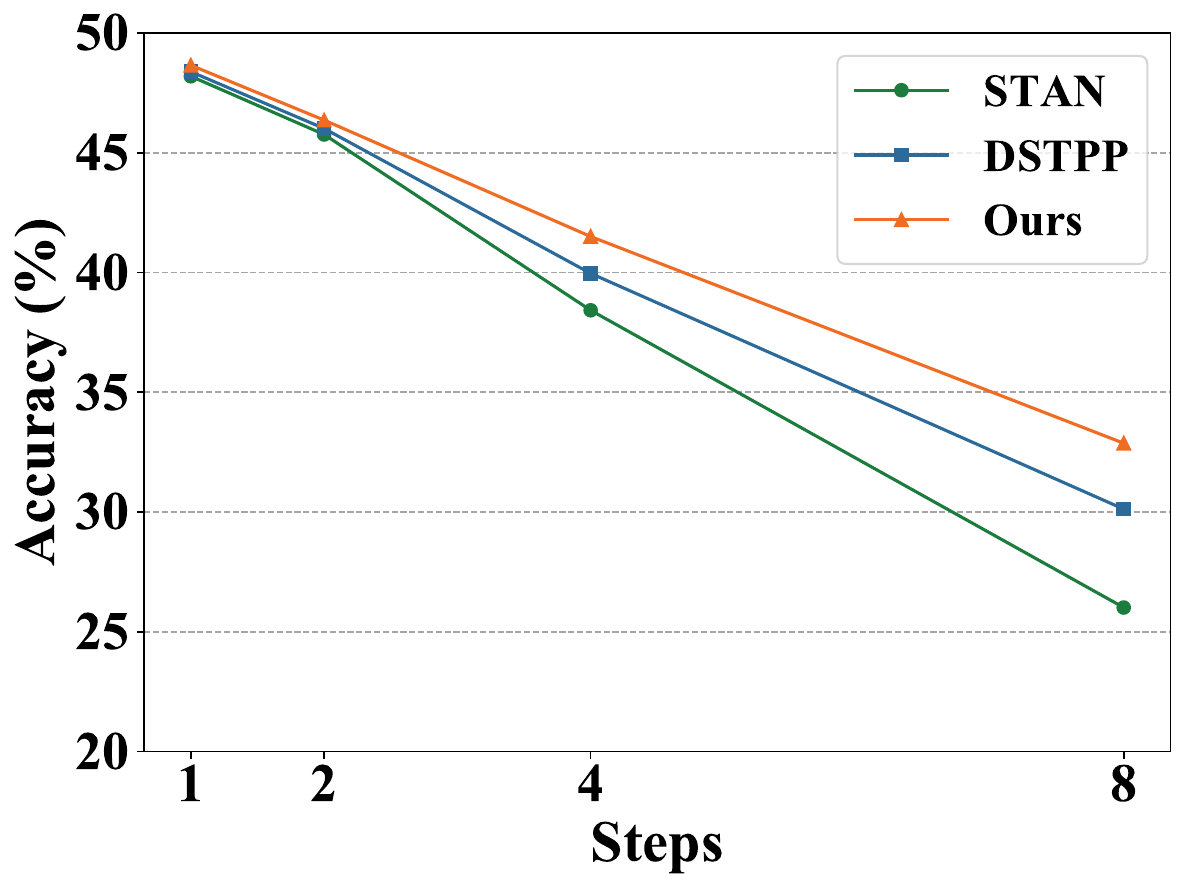}}
\hfil
\subfloat[Long-term prediction on ISP dataset]{\includegraphics[width=.23\textwidth]{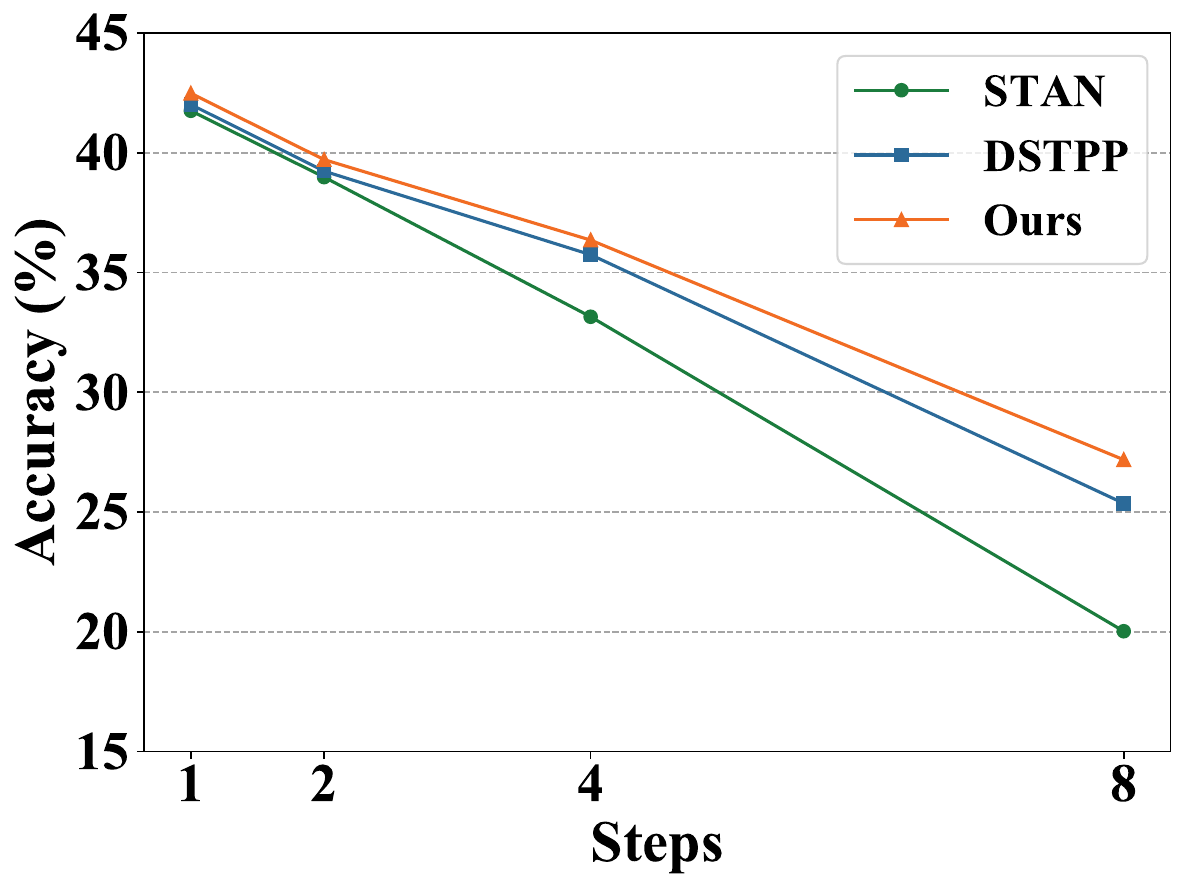}}
\caption{Performance of long-term trajectory prediction on two datasets.} 
\label{fig:long_prediction}
\end{figure}

\subsubsection{Trajectory Prediction}
The performance of our method on trajectory prediction is shown in Table~\ref{tab:Next Location Prediction comparison}. For this task, our performance is only close to the best baseline, possibly because our model loses its specific design for single tasks to ensure generality for all tasks.
However, our model significantly outperforms other baseline models for extended task long-term trajectory prediction. Figure~\ref{fig:long_prediction} shows the performance of our model in long-term trajectory prediction, predicting up to four hours (8 steps, with each step representing half an hour). We can observe that the performance of all models decreases as the prediction time step increases. However, our model consistently shows the best performance across different time steps. This indicates that our model is more suitable for long-term trajectory prediction tasks.

\subsubsection{Trajectory Recovery}
The performance of our method on trajectory recovery is shown in Table~\ref{tab:Trajectory Recovery comparison}. 
Our model shows a significant performance improvement on the most representative metric Distance, achieving around 6\% improvement.
In the real world, the collected trajectory data is often sparse due to the low sampling frequency and limited coverage of the collection devices. Therefore, scarcity-constrained trajectory prediction is also a practical task. Since the current baseline cannot complete this task independently, we choose the best trajectory recovery baseline to convert sparse trajectories into dense ones and then use a trajectory prediction model for the next location prediction. As shown in Figure~\ref{fig:incomplete-prediction}, our model exhibits an average performance improvement of 6.2\% over the best baseline, because our random masking strategy allows the model to be trained with more trajectory samples with scarce history, enabling it to handle this task.

\subsection{Zero-shot Performance}
Zero-shot evaluation can demonstrate the adaptability of models. In datasets segmented by users, we train GenMove with data from some users and then evaluate it on a completely unseen user dataset. Since the objective of the generation task is to model the trajectory distribution of all users without distinguishing between individuals, we do not create a new user-based dataset for generation tasks. We only performed zero-shot inference for tasks such as trajectory prediction and recovery.

\begin{figure}[t]
\centering
\subfloat[Trajectory recovey]{\includegraphics[height=.170\textwidth]{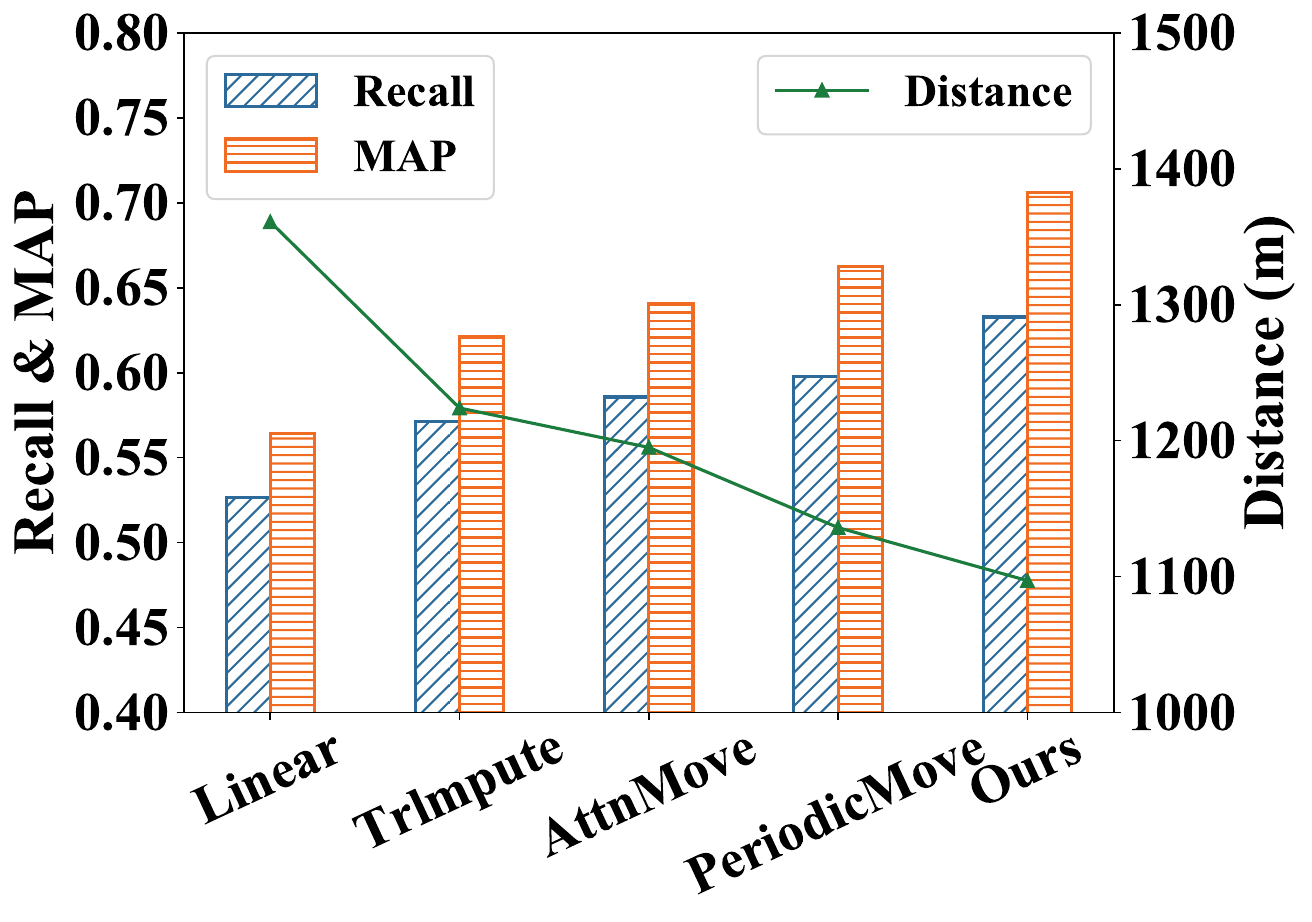}}
\hfil
\subfloat[Trajectory prediction]{\includegraphics[height=.170\textwidth]{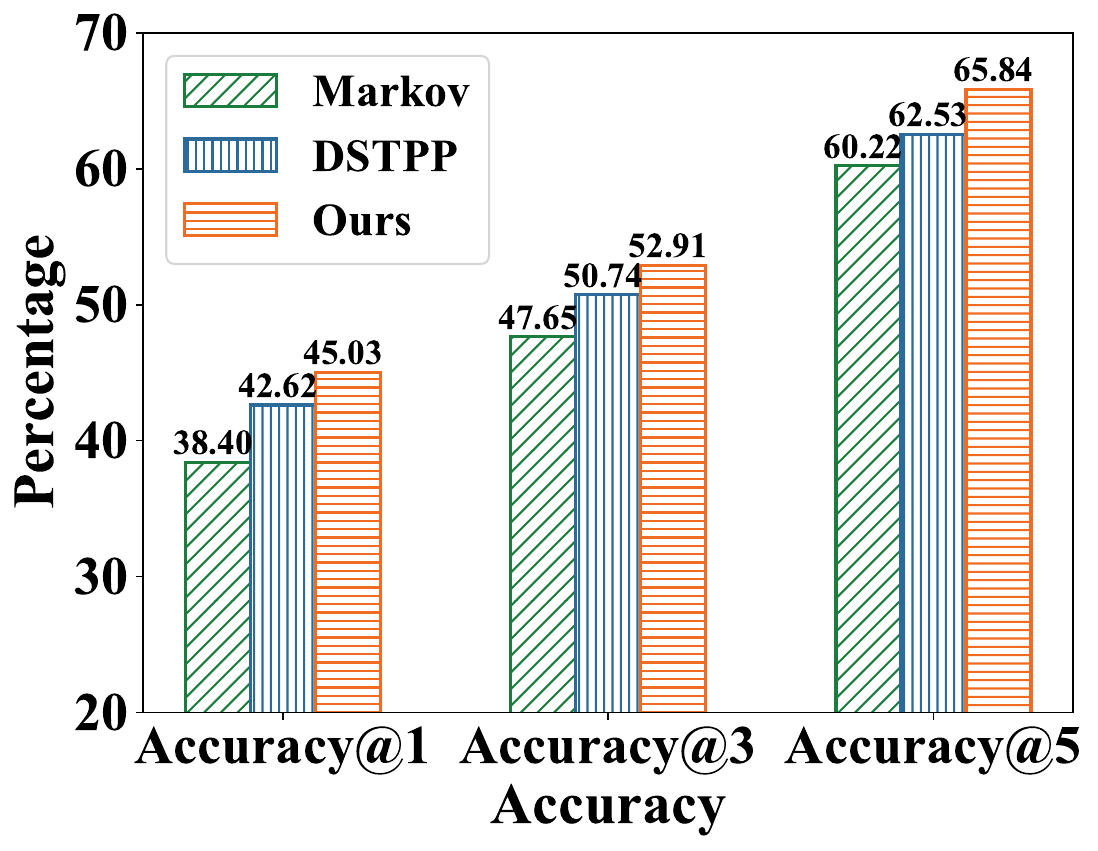}}
\hfil
\subfloat[Scarcity-constrained prediction]{\includegraphics[width=.23\textwidth]{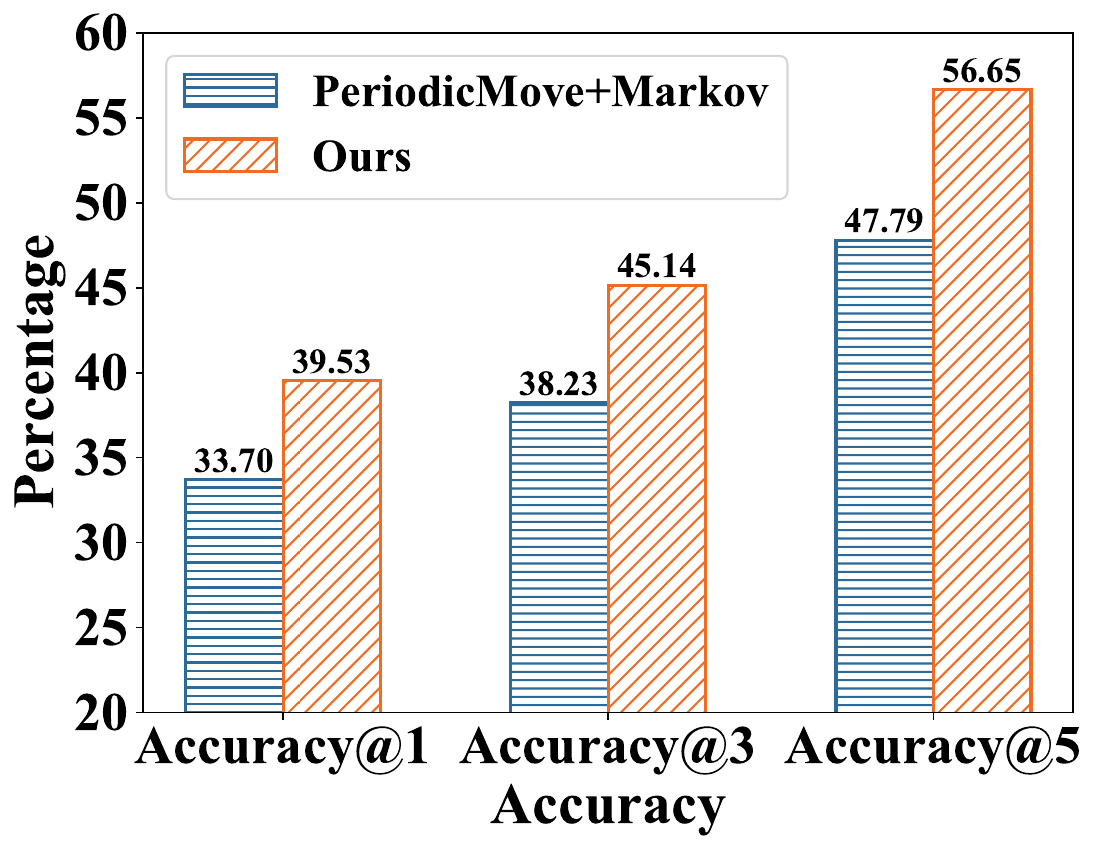}}
\hfil
\subfloat[Long-term prediction]{\includegraphics[width=.23\textwidth]{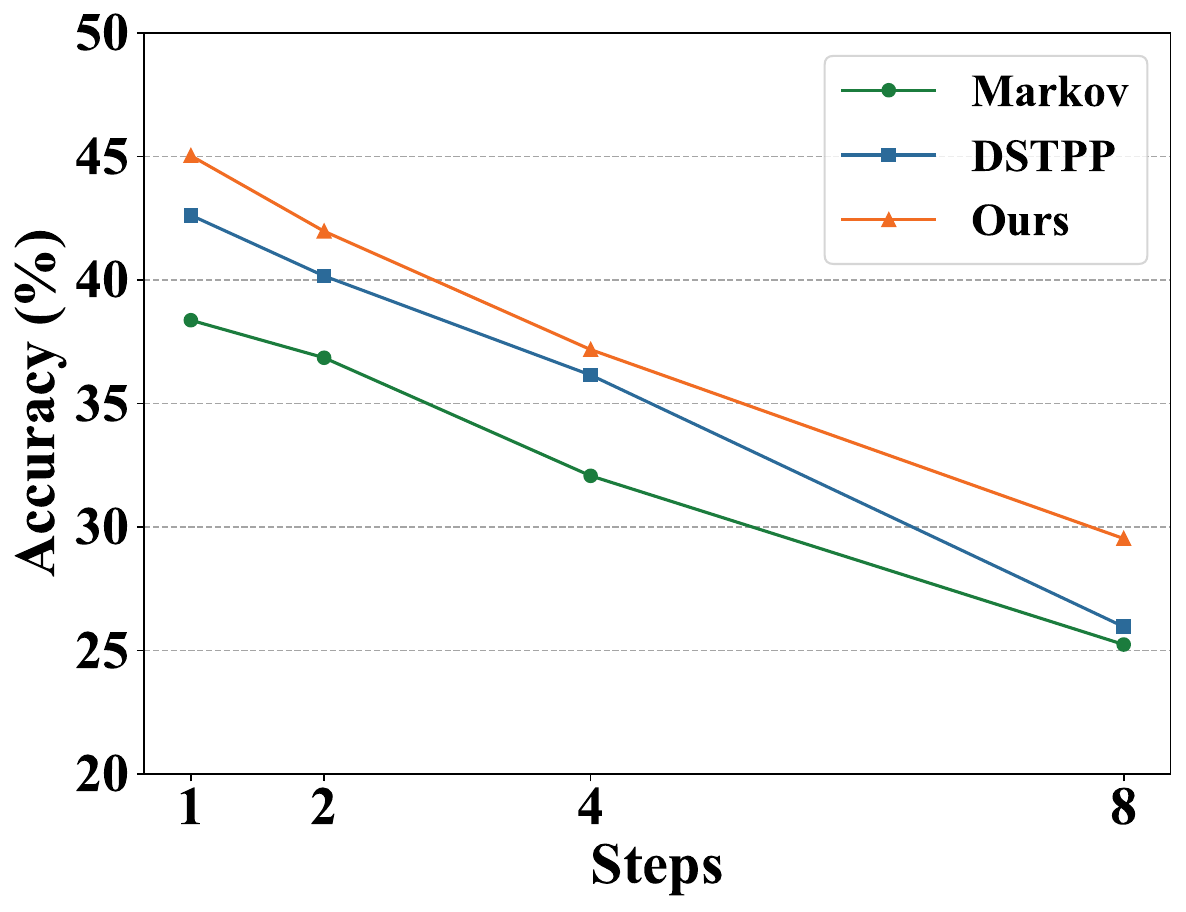}}
\caption{Zero-shot performance in four tasks on the MME dataset.} 
\label{fig:user}
\end{figure}

As shown in Figure~\ref{fig:user}, GenMove exhibits exceptional zero-shot performance, significantly surpassing the best baseline. In scarcity-constrained prediction, it improves the Accuracy@5 metric by more than 18\% compared to the best baseline; in long-term prediction tasks, the performance of predicting up to 8 steps ahead has improved by nearly 20\%. These remarkable results demonstrate GenMove's ability to extract and learn similar mobility patterns from historical data. The zero-shot results further verify GenMove's strong generalization capabilities in entirely new scenarios.

\begin{figure}[t]
\centering
\subfloat[Trajectory recovery on MME dataset]{\includegraphics[width=.23\textwidth]{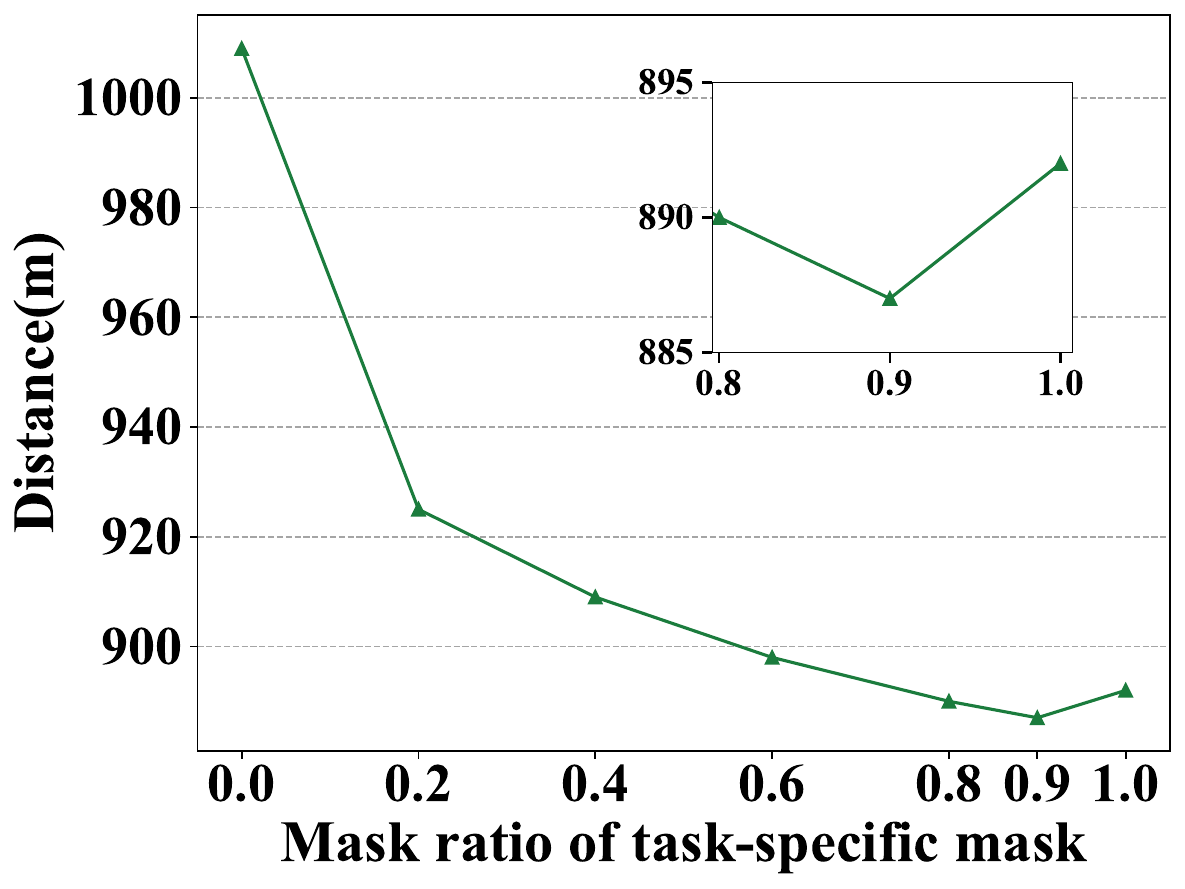}}
\hfil
\subfloat[Trajectory prediction on MME dataset]{\includegraphics[width=.23\textwidth]
{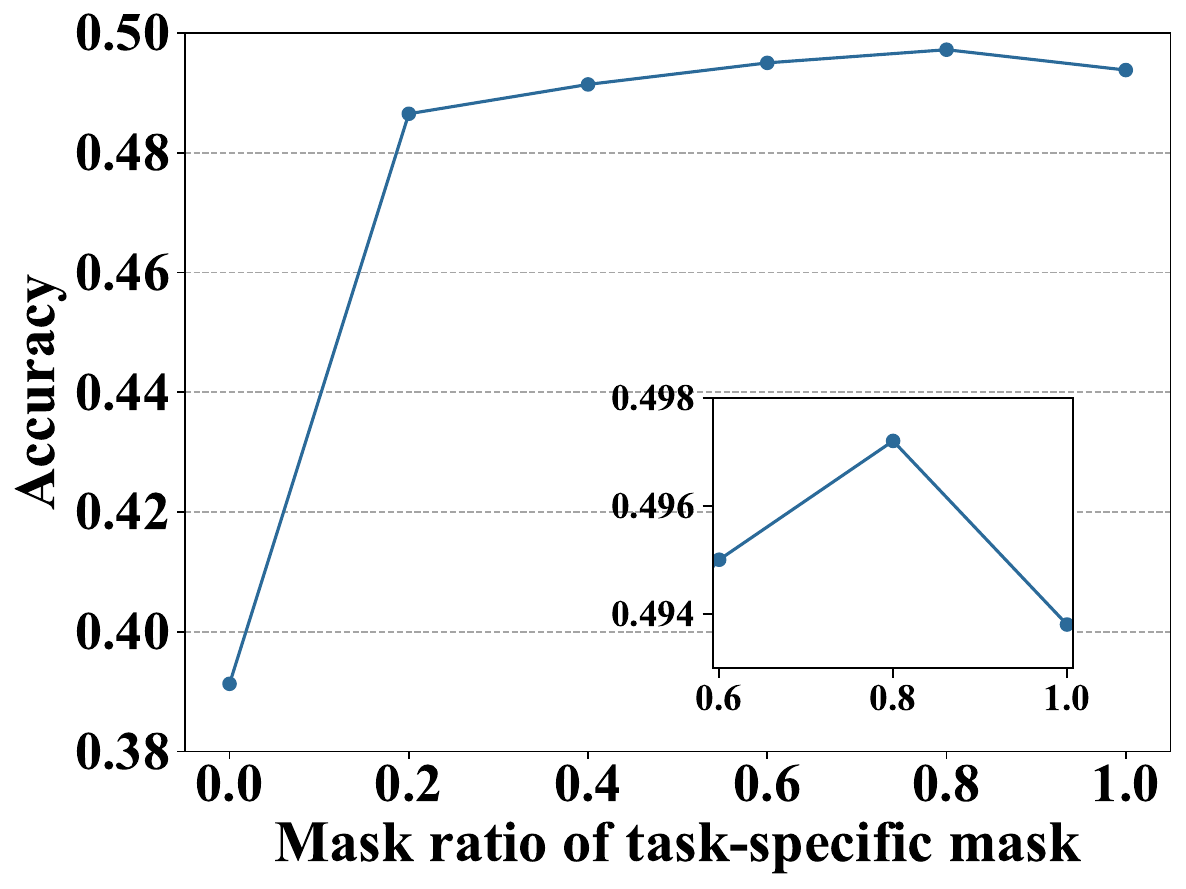}}
\hfil
\subfloat[Trajectory recovery on ISP dataset]{\includegraphics[width=.23\textwidth]{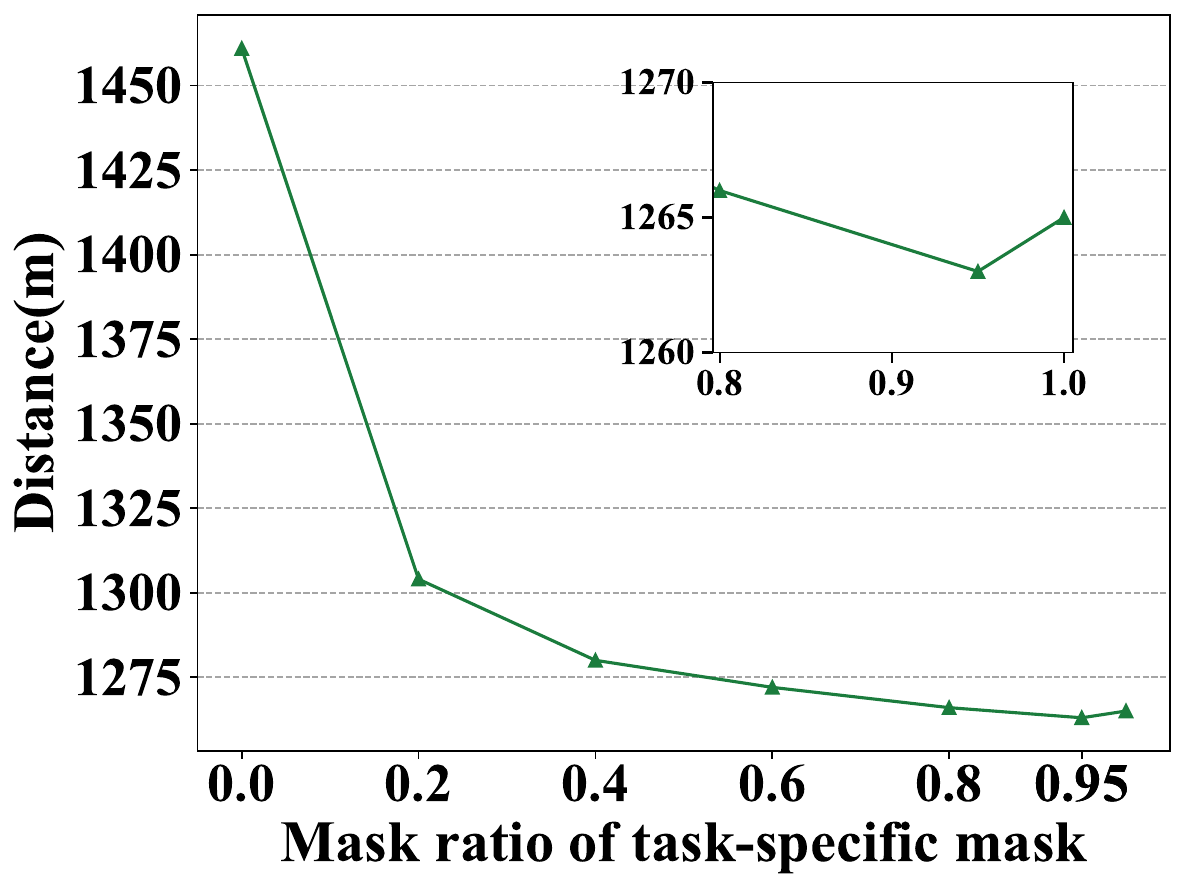}}
\hfil
\subfloat[Trajectory prediction on ISP dataset]{\includegraphics[width=.23\textwidth]{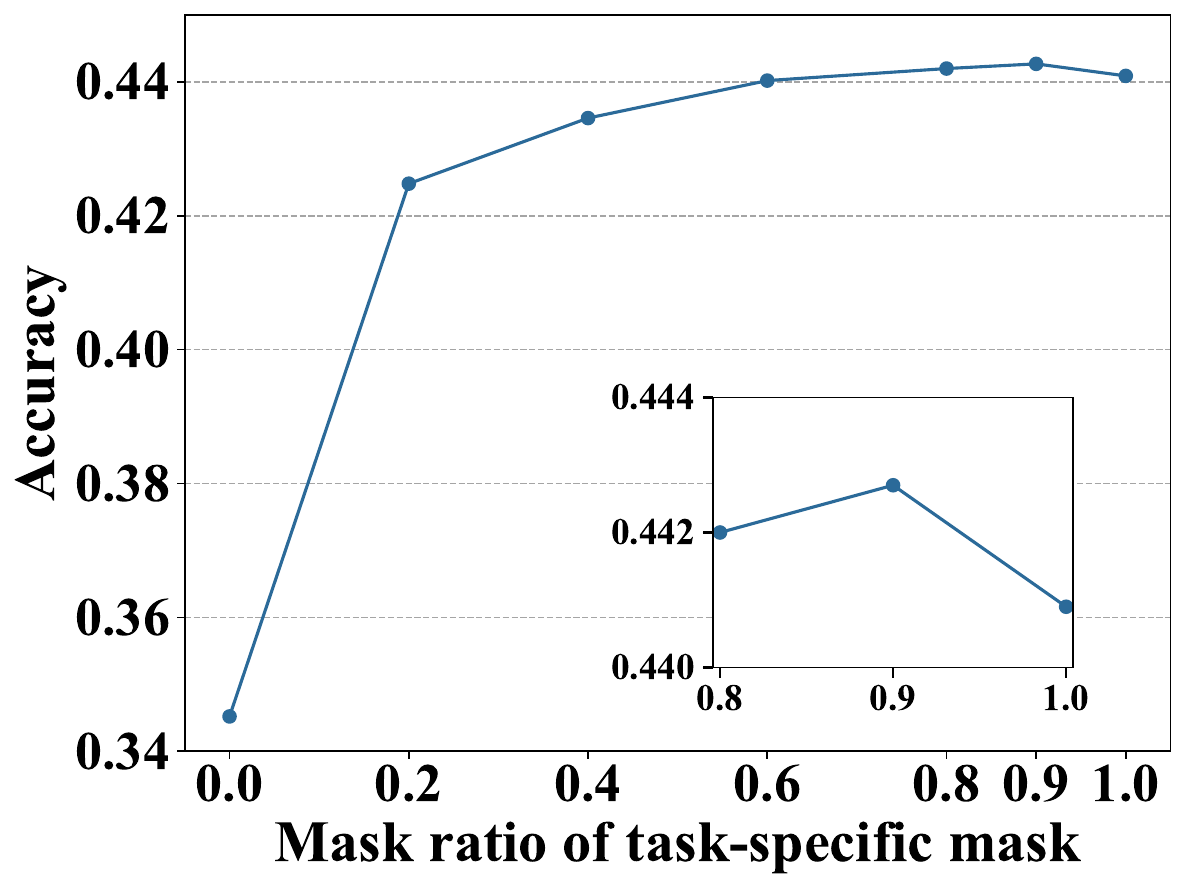}}
\caption{Impact of the mask ratio on the performance of different tasks on two datasets.} 
\label{fig:ratio}
\end{figure}

\subsection{Mutual Enhancement between Tasks}
We analyze the relationship between mask ratio and task performance to explore whether frameworks that train multiple tasks together have advantages over traditional single task. 


As shown in Figure~\ref{fig:ratio}, when the recovery or prediction task-specific mask (i.e., random masks) ratio is particularly small, the performance of the recovery or prediction task is poor. This proves the effectiveness of the task-specific masks. Moreover, we find an interesting phenomenon: the optimized performance is not at a mask ratio of 1, indicating that the general framework performs better when trained on multiple tasks simultaneously than when trained on single tasks separately. 
For example, Figure \ref{fig:ratio} (b) indicates optimal performance is at a mask ratio of 0.8 rather than 1, suggesting that the recovery or generation task enhances the prediction task. This supports our general framework’s premise that different tasks can boost performance through shared patterns.
This demonstrates that the general framework performs better when trained on multiple tasks simultaneously than when trained on single tasks separately. It further illustrates the motivation behind our general framework that different tasks can improve performance by sharing mobility patterns.

\section{Conclusion}
In this paper, we address an important problem of building a general framework that accomplishes multiple tasks for mobility trajectory modeling. 
Diverse formats of different tasks are unified into a standard format of task objectives and conditional observations through masking strategies. 
By introducing contextual trajectory embedding through a classifier-free guidance approach, our model can flexibly adapt to different conditions.
Extensive experiments conducted on diverse datasets with multiple evaluation metrics have demonstrated the superiority of our model. 
In the future, we plan to adopt transfer learning techniques to achieve mobility modeling across different cities, further enhancing the generality of our framework. Moreover, we can use different backbones to explore their impact on performance or generality.

%

\bibliographystyle{IEEEtran}

\bibliography{sample-base}

\begin{thebibliography}{10}
\providecommand{\url}[1]{#1}
\csname url@samestyle\endcsname
\providecommand{\newblock}{\relax}
\providecommand{\bibinfo}[2]{#2}
\providecommand{\BIBentrySTDinterwordspacing}{\spaceskip=0pt\relax}
\providecommand{\BIBentryALTinterwordstretchfactor}{4}
\providecommand{\BIBentryALTinterwordspacing}{\spaceskip=\fontdimen2\font plus
\BIBentryALTinterwordstretchfactor\fontdimen3\font minus \fontdimen4\font\relax}
\providecommand{\BIBforeignlanguage}[2]{{%
\expandafter\ifx\csname l@#1\endcsname\relax
\typeout{** WARNING: IEEEtran.bst: No hyphenation pattern has been}%
\typeout{** loaded for the language `#1'. Using the pattern for}%
\typeout{** the default language instead.}%
\else
\language=\csname l@#1\endcsname
\fi
#2}}
\providecommand{\BIBdecl}{\relax}
\BIBdecl

\bibitem{li2025generative}
T.~Li, Q.~Long, H.~Chai, S.~Zhang, F.~Jiang, H.~Liu, W.~Huang, D.~Jin, and Y.~Li, ``Generative ai empowered network digital twins: Architecture, technologies, and applications,'' \emph{ACM Computing Surveys}, 2025.

\bibitem{wang2021public}
S.~Wang, Y.~Sun, C.~Musco, and Z.~Bao, ``Public transport planning: When transit network connectivity meets commuting demand,'' in \emph{Proceedings of the 2021 International Conference on Management of Data}, 2021, pp. 1906--1919.

\bibitem{yabe2024enhancing}
T.~Yabe, M.~Luca, K.~Tsubouchi, B.~Lepri, M.~C. Gonzalez, and E.~Moro, ``Enhancing human mobility research with open and standardized datasets,'' \emph{Nature Computational Science}, pp. 1--4, 2024.

\bibitem{feng2018deepmove}
J.~Feng, Y.~Li, C.~Zhang, F.~Sun, F.~Meng, A.~Guo, and D.~Jin, ``Deepmove: Predicting human mobility with attentional recurrent networks,'' in \emph{Proceedings of the 2018 world wide web conference}, 2018, pp. 1459--1468.

\bibitem{long2024universal}
Q.~Long, Y.~Yuan, and Y.~Li, ``A universal model for human mobility prediction,'' \emph{arXiv preprint arXiv:2412.15294}, 2024.

\bibitem{jiang2016timegeo}
S.~Jiang, Y.~Yang, S.~Gupta, D.~Veneziano, S.~Athavale, and M.~C. Gonz{\'a}lez, ``The timegeo modeling framework for urban mobility without travel surveys,'' \emph{PNAS}, vol. 113, no.~37, 2016.

\bibitem{cao2021generating}
C.~Cao and M.~Li, ``Generating mobility trajectories with retained data utility,'' in \emph{Proceedings of the 27th ACM SIGKDD Conference on Knowledge Discovery \& Data Mining}, 2021, pp. 2610--2620.

\bibitem{elshrif2022network}
M.~M. Elshrif, K.~Isufaj, and M.~F. Mokbel, ``Network-less trajectory imputation,'' in \emph{Proceedings of the 30th International Conference on Advances in Geographic Information Systems}, 2022, pp. 1--10.

\bibitem{xia2021attnmove}
T.~Xia, Y.~Qi, J.~Feng, F.~Xu, F.~Sun, D.~Guo, and Y.~Li, ``Attnmove: History enhanced trajectory recovery via attentional network,'' in \emph{Proceedings of the AAAI Conference on Artificial Intelligence}, vol.~35, no.~5, 2021, pp. 4494--4502.

\bibitem{xu2024representation}
X.~Xu, C.~Yang, and W.~Wu, ``Representation learning and graph convolutional networks for short-term vehicle trajectory prediction,'' \emph{Physica A: Statistical Mechanics and its Applications}, vol. 637, p. 129560, 2024.

\bibitem{rossi2021vehicle}
L.~Rossi, A.~Ajmar, M.~Paolanti, and R.~Pierdicca, ``Vehicle trajectory prediction and generation using lstm models and gans,'' \emph{Plos one}, vol.~16, no.~7, p. e0253868, 2021.

\bibitem{chen2024advancements}
H.~Chen, H.~Wang, Q.~Long, D.~Jin, and Y.~Li, ``Advancements in federated learning: Models, methods, and privacy,'' \emph{ACM Computing Surveys}, vol.~57, no.~2, pp. 1--39, 2024.

\bibitem{zhu2023diffusion}
Y.~Zhu, Y.~Ye, X.~Zhao, and J.~J. Yu, ``Diffusion model for gps trajectory generation,'' \emph{arXiv preprint arXiv:2304.11582}, 2023.

\bibitem{wang2023pategail}
H.~Wang, C.~Gao, Y.~Wu, D.~Jin, L.~Yao, and Y.~Li, ``Pategail: A privacy-preserving mobility trajectory generator with imitation learning,'' in \emph{Proceedings of the AAAI Conference on Artificial Intelligence}, vol.~37, no.~12, 2023, pp. 14\,539--14\,547.

\bibitem{dai2015personalized}
J.~Dai, B.~Yang, C.~Guo, and Z.~Ding, ``Personalized route recommendation using big trajectory data,'' in \emph{2015 IEEE 31st international conference on data engineering}.\hskip 1em plus 0.5em minus 0.4em\relax IEEE, 2015, pp. 543--554.

\bibitem{xing2017personalized}
L.~G. Xing, I.~A. Abiodun, C.~W. Khuen, and T.~T. Boon, ``A personalized recommendation framework with user trajectory analysis applied in location-based social network (lbsn),'' in \emph{2017 IEEE 3rd International Conference on Engineering Technologies and Social Sciences (ICETSS)}.\hskip 1em plus 0.5em minus 0.4em\relax IEEE, 2017, pp. 1--6.

\bibitem{long2022vision}
S.~Long, F.~Cao, S.~C. Han, and H.~Yang, ``Vision-and-language pretrained models: A survey,'' \emph{arXiv preprint arXiv:2204.07356}, 2022.

\bibitem{xie2023towards}
L.~Xie, L.~Wei, X.~Zhang, K.~Bi, X.~Gu, J.~Chang, and Q.~Tian, ``Towards agi in computer vision: Lessons learned from gpt and large language models,'' \emph{arXiv preprint arXiv:2306.08641}, 2023.

\bibitem{achiam2023gpt}
J.~Achiam, S.~Adler, S.~Agarwal, L.~Ahmad, I.~Akkaya, F.~L. Aleman, D.~Almeida, J.~Altenschmidt, S.~Altman, S.~Anadkat \emph{et~al.}, ``Gpt-4 technical report,'' \emph{arXiv preprint arXiv:2303.08774}, 2023.

\bibitem{feng2020learning}
J.~Feng, Z.~Yang, F.~Xu, H.~Yu, M.~Wang, and Y.~Li, ``Learning to simulate human mobility,'' in \emph{Proc. KDD}, 2020.

\bibitem{sun2021periodicmove}
H.~Sun, C.~Yang, L.~Deng, F.~Zhou, F.~Huang, and K.~Zheng, ``Periodicmove: shift-aware human mobility recovery with graph neural network,'' in \emph{Proceedings of the 30th ACM International Conference on Information \& Knowledge Management}, 2021, pp. 1734--1743.

\bibitem{qingyue2024privacy}
L.~Qingyue, W.~Huandong, C.~Huiming, J.~Depeng, Z.~Lin, Y.~Li, and L.~Yong, ``Privacy-preserving federated mobility prediction with compound data and model perturbation mechanism,'' \emph{China Communications}, vol.~21, no.~3, pp. 160--173, 2024.

\bibitem{song2010modelling}
C.~Song, T.~Koren, P.~Wang, and A.-L. Barabási, ``Modelling the scaling properties of human mobility,'' \emph{Nature Physics}, vol.~6, no.~10, pp. 818--823, 2010.

\bibitem{xue2021mobtcast}
H.~Xue, F.~Salim, Y.~Ren, and N.~Oliver, ``Mobtcast: Leveraging auxiliary trajectory forecasting for human mobility prediction,'' \emph{Advances in Neural Information Processing Systems}, vol.~34, pp. 30\,380--30\,391, 2021.

\bibitem{luo2021stan}
Y.~Luo, Q.~Liu, and Z.~Liu, ``Stan: Spatio-temporal attention network for next location recommendation,'' in \emph{Proceedings of the web conference 2021}, 2021, pp. 2177--2185.

\bibitem{liu2024sora}
Y.~Liu, K.~Zhang, Y.~Li, Z.~Yan, C.~Gao, R.~Chen, Z.~Yuan, Y.~Huang, H.~Sun, J.~Gao \emph{et~al.}, ``Sora: A review on background, technology, limitations, and opportunities of large vision models,'' \emph{arXiv preprint arXiv:2402.17177}, 2024.

\bibitem{garza2023timegpt}
A.~Garza and M.~Mergenthaler-Canseco, ``Timegpt-1,'' \emph{arXiv preprint arXiv:2310.03589}, 2023.

\bibitem{wu2022timesnet}
H.~Wu, T.~Hu, Y.~Liu, H.~Zhou, J.~Wang, and M.~Long, ``Timesnet: Temporal 2d-variation modeling for general time series analysis,'' \emph{arXiv preprint arXiv:2210.02186}, 2022.

\bibitem{chu2024simulating}
C.~Chu, H.~Zhang, P.~Wang, and F.~Lu, ``Simulating human mobility with a trajectory generation framework based on diffusion model,'' \emph{International Journal of Geographical Information Science}, pp. 1--32, 2024.

\bibitem{jin2023time}
M.~Jin, S.~Wang, L.~Ma, Z.~Chu, J.~Y. Zhang, X.~Shi, P.-Y. Chen, Y.~Liang, Y.-F. Li, S.~Pan \emph{et~al.}, ``Time-llm: Time series forecasting by reprogramming large language models,'' \emph{arXiv preprint arXiv:2310.01728}, 2023.

\bibitem{li2024urbangpt}
Z.~Li, L.~Xia, J.~Tang, Y.~Xu, L.~Shi, L.~Xia, D.~Yin, and C.~Huang, ``Urbangpt: Spatio-temporal large language models,'' \emph{arXiv preprint arXiv:2403.00813}, 2024.

\bibitem{tashiro2021csdi}
Y.~Tashiro, J.~Song, Y.~Song, and S.~Ermon, ``Csdi: Conditional score-based diffusion models for probabilistic time series imputation,'' \emph{Advances in Neural Information Processing Systems}, vol.~34, pp. 24\,804--24\,816, 2021.

\bibitem{ho2020denoising}
J.~Ho, A.~Jain, and P.~Abbeel, ``Denoising diffusion probabilistic models,'' \emph{Advances in Neural Information Processing Systems}, vol.~33, pp. 6840--6851, 2020.

\bibitem{yuan2023spatio}
Y.~Yuan, J.~Ding, C.~Shao, D.~Jin, and Y.~Li, ``Spatio-temporal diffusion point processes,'' \emph{arXiv preprint arXiv:2305.12403}, 2023.

\bibitem{zhou2023towards}
Z.~Zhou, J.~Ding, Y.~Liu, D.~Jin, and Y.~Li, ``Towards generative modeling of urban flow through knowledge-enhanced denoising diffusion,'' in \emph{Proceedings of the 31st ACM International Conference on Advances in Geographic Information Systems}, 2023, pp. 1--12.

\bibitem{tang2015line}
J.~Tang, M.~Qu, M.~Wang, M.~Zhang, J.~Yan, and Q.~Mei, ``Line: Large-scale information network embedding,'' in \emph{Proceedings of the 24th international conference on world wide web}, 2015, pp. 1067--1077.

\bibitem{liu2023summary}
Y.~Liu, T.~Han, S.~Ma, J.~Zhang, Y.~Yang, J.~Tian, H.~He, A.~Li, M.~He, Z.~Liu \emph{et~al.}, ``Summary of chatgpt-related research and perspective towards the future of large language models,'' \emph{Meta-Radiology}, p. 100017, 2023.

\bibitem{zhang2022human}
B.~Zhang, T.~Wang, C.~Zhou, N.~Conci, and H.~Liu, ``Human trajectory forecasting using a flow-based generative model,'' \emph{Engineering Applications of Artificial Intelligence}, vol. 115, p. 105236, 2022.

\bibitem{vaswani2017attention}
A.~Vaswani, N.~Shazeer, N.~Parmar, J.~Uszkoreit, L.~Jones, A.~N. Gomez, {\L}.~Kaiser, and I.~Polosukhin, ``Attention is all you need,'' \emph{Advances in neural information processing systems}, vol.~30, 2017.

\bibitem{kong2020diffwave}
Z.~Kong, W.~Ping, J.~Huang, K.~Zhao, and B.~Catanzaro, ``Diffwave: A versatile diffusion model for audio synthesis,'' \emph{arXiv preprint arXiv:2009.09761}, 2020.

\bibitem{ho2022classifier}
J.~Ho and T.~Salimans, ``Classifier-free diffusion guidance,'' \emph{arXiv preprint arXiv:2207.12598}, 2022.

\bibitem{torabi2018behavioral}
F.~Torabi, G.~Warnell, and P.~Stone, ``Behavioral cloning from observation,'' \emph{arXiv preprint arXiv:1805.01954}, 2018.

\bibitem{long2023practical}
Q.~Long, H.~Wang, T.~Li, L.~Huang, K.~Wang, Q.~Wu, G.~Li, Y.~Liang, L.~Yu, and Y.~Li, ``Practical synthetic human trajectories generation based on variational point processes,'' in \emph{Proceedings of the 29th ACM SIGKDD Conference on Knowledge Discovery and Data Mining}, 2023, pp. 4561--4571.

\bibitem{hoteit2014estimating}
S.~Hoteit, S.~Secci, S.~Sobolevsky, C.~Ratti, and G.~Pujolle, ``Estimating human trajectories and hotspots through mobile phone data,'' \emph{Computer Networks}, vol.~64, pp. 296--307, 2014.

\bibitem{li2019reconstruction}
M.~Li, S.~Gao, F.~Lu, and H.~Zhang, ``Reconstruction of human movement trajectories from large-scale low-frequency mobile phone data,'' \emph{Computers, Environment and Urban Systems}, vol.~77, p. 101346, 2019.

\bibitem{gambs2012next}
S.~Gambs, M.-O. Killijian, and M.~N. del Prado~Cortez, ``Next place prediction using mobility markov chains,'' in \emph{Proceedings of the first workshop on measurement, privacy, and mobility}, 2012, pp. 1--6.

\bibitem{Kong2018HST}
D.~Kong and F.~Wu, ``Hst-lstm: A hierarchical spatial-temporal long-short term memory network for location prediction,'' in \emph{Twenty-Seventh International Joint Conference on Artificial Intelligence {IJCAI-18}}, 2018.

\end{thebibliography}


\vfill

\end{document}